\documentclass{article}

\usepackage{arxiv}

\usepackage[utf8]{inputenc} 
\usepackage[T1]{fontenc}    
\usepackage{hyperref}       
\usepackage{url}            
\usepackage{booktabs}       
\usepackage{amsfonts}       
\usepackage{nicefrac}       
\usepackage{microtype}      
\usepackage{lipsum}		
\usepackage{graphicx}
\usepackage{doi}
\usepackage{amsmath}
\usepackage{tikz}
\usepackage{varwidth}
\usepackage[numbers]{natbib}

\title{Resource-Efficient Deep Learning: A Survey on Model-, Arithmetic-, and Implementation-Level Techniques}

\date{} 					

\author{JunKyu Lee \\
	Queen's University Belfast\\
	Belfast, Northern Ireland, UK\\
	\texttt{junkyu.lee@qub.ac.uk} \\
	\And
Lev Mukhanov \\
	Queen's University Belfast\\
	Belfast, Northern Ireland, UK\\
	\texttt{l.mukhanov@qub.ac.uk} \\
	\And
Amir Sabbagh Molahosseini \\
	Queen's University Belfast\\
	Belfast, Northern Ireland, UK\\
	\texttt{a.sabbaghmolahosseini@qub.ac.uk} \\
		\And
Umar Minhas \\
	Queen's University Belfast\\
	Belfast, Northern Ireland, UK\\
	\texttt{u.minhas@qub.ac.uk} \\
		\And
Yang Hua \\
	Queen's University Belfast\\
	Belfast, Northern Ireland, UK\\
	\texttt{y.hua@qub.ac.uk} \\
		\And
Jesus Martinez del Rincon \\
	Queen's University Belfast\\
	Belfast, Northern Ireland, UK\\
	\texttt{j.martinez-del-rincon@qub.ac.uk} \\
		\And
Kiril Dichev\\
	University of Cambridge\\
	Cambridge, England, UK\\
	\texttt{kiril.dichev@gmail.com} 
	\And
Cheol-Ho Hong \thanks{Corresponding Author} \\
	Chung-Ang University\\
	Seoul, South Korea\\
	\texttt{cheolhohong@cau.ac.kr} \\
		\And
Hans Vandierendonck \\
	Queen's University Belfast\\
	Belfast, Northern Ireland, UK\\
	\texttt{h.vandierendonck@qub.ac.uk}  }

\renewcommand{\shorttitle}{\textit{arXiv} Template}

\hypersetup{
pdftitle={A template for the arxiv style},
pdfsubject={q-bio.NC, q-bio.QM},
pdfauthor={David S.~Hippocampus, Elias D.~Striatum},
pdfkeywords={First keyword, Second keyword, More},
}

\begin{document}
\maketitle

\begin{abstract}
Deep learning is pervasive in our daily life, including self-driving cars, virtual assistants, social network services, healthcare services, face recognition, etc. However, deep neural networks demand substantial compute resources during training and inference. The machine learning community has mainly focused on model-level optimizations such as architectural compression of deep learning models, while the system community has focused on implementation-level optimization. In between, various arithmetic-level optimization techniques have been proposed in the arithmetic community. This article provides a survey on resource-efficient deep learning techniques in terms of model-, arithmetic-, and  implementation-level techniques and identifies the research gaps for resource-efficient deep learning techniques across the three different level techniques. Our survey clarifies the influence from higher to lower-level techniques based on our resource-efficiency metric definition and discusses the future trend for resource-efficient deep learning research.
\end{abstract}


\section{Introduction}
Recent improvements in network and storage devices have provided the machine learning community  with the opportunity to utilize immense data sources, leading to the golden age of AI and deep learning \cite{bottou}. Since modern deep neural networks (DNNs) require considerable computing resources and are deployed in a variety of compute devices, ranging from high-end servers to mobile devices with limited computational resources, there is a strong need to realize economical DNNs that fit within the resource constraints~\cite{panda-conditional, teer-branchy, teer-distributed}. Resource-efficient deep learning research has vividly been carried out independently in various research communities including the machine learning, computer arithmetic, and computing system communities. Recently, DeepMind proposed the resource-efficient deep learning benchmark metric which is the accuracy along with the required memory footprint and number of operations \cite{hu-one-pass}.  

\begin{figure}[!t] 
\centering
\includegraphics[width=4in]{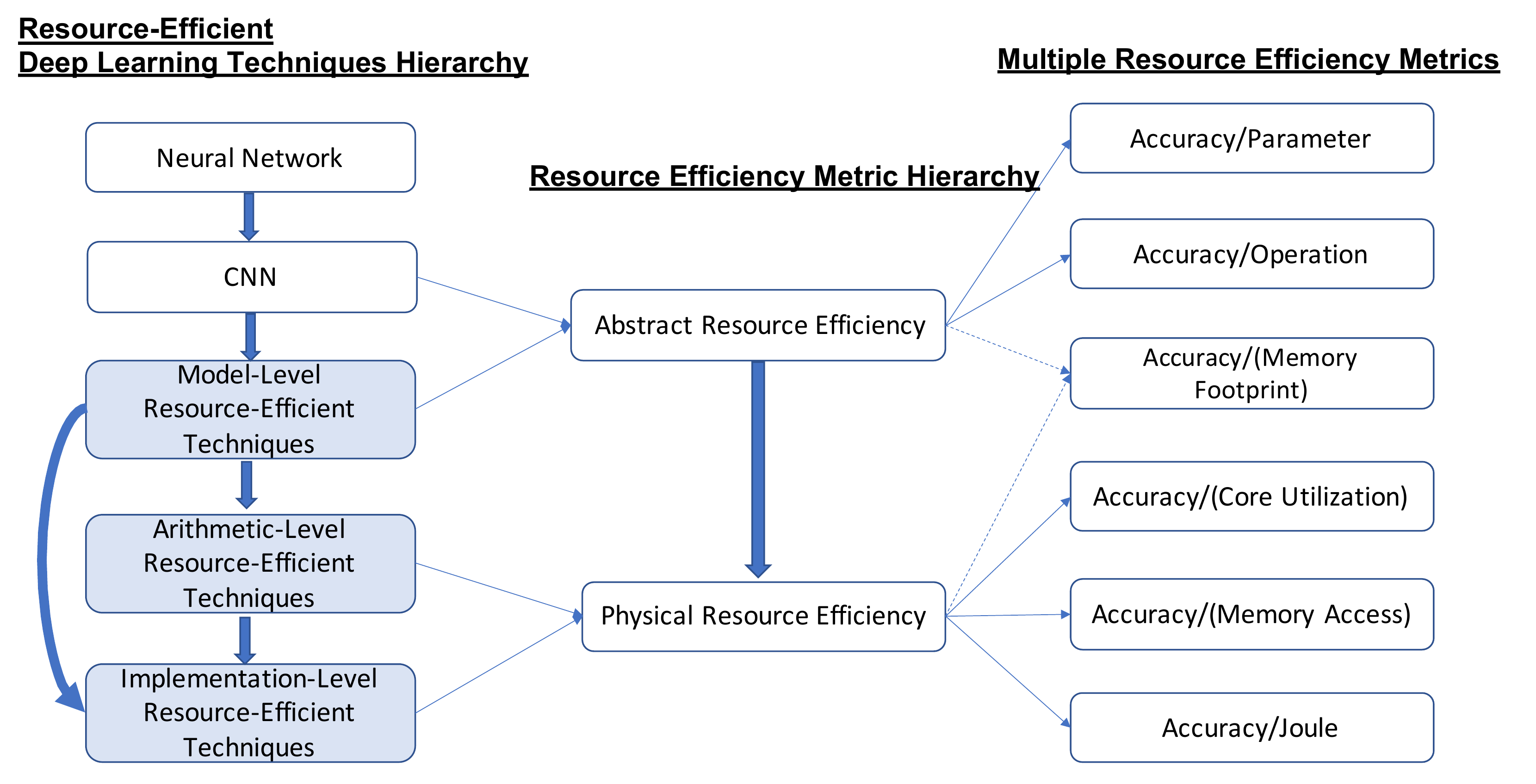}
\caption{Survey on resource-efficient deep learning techniques based on resource efficiency metrics.}
\label{fig:survey-diagram}
\end{figure}

With this regard, this article surveys resource-efficient techniques for deep learning based on the three-level categorization: the model-, arithmetic-, and implementation-level techniques along with various resource efficiency metrics as shown in Fig.~\ref{fig:survey-diagram}. Our resource efficiency metrics include the accuracy per parameter, operation, memory footprint, core utilization, memory access, and Joule. For the resource-efficiency comparison between the baseline DNN and a DNN utilizing resource-efficient techniques, the accuracy should be equivalent between the two DNNs. In other words, it is not fair to compare the resource efficiency between a DNN producing a high accuracy and a DNN producing a low accuracy since the resource efficiency is significantly higher in a low performing DNN based on our resource metrics. We categorize the resource-efficient techniques into the \textit{model-level resource-efficient techniques} if they compress the DNN model sizes; the \textit{arithmetic-level resource-efficient techniques} if they utilize reduced precision arithmetic and/or customized arithmetic rules; the \textit{implementation-level resource-efficient techniques} if they apply hardware optimization techniques to the DNNs (e.g., locating local memory near to processing elements) to improve physical resource efficiency such as the accuracy per compute resource and per Joule. 

In Fig.~\ref{fig:survey-diagram}, Convolutional Neural Networks (CNNs) can be considered as a resource-efficient technique since they improve the accuracy per parameter, per operation, and per memory footprint, compared to fully connected neural networks. The resource-efficiency from CNNs can be further improved by applying the model-, arithmetic-, and implementation-level techniques. The model- and arithmetic-level techniques can affect the accuracy since they affect either the DNN model structure or the arithmetic rule, while the implementation-level techniques generally do not affect the accuracy. The model-level techniques mostly contribute to improving abstract resource efficiency, while the implementation-level techniques contribute to improve physical resource efficiency. Without careful consideration at the intersection between the model- and the implementation-level techniques, a DNN model compressed by the model-level techniques might require significant runtime compute resources, incurring longer training time and inference latency than the original model~\cite{ma-shufflenetv2, chen-eyeriss}. Thus, to optimize the performance and energy efficiency on a particular hardware, it is essential to consider the joint effect of the model, arithmetic and implementation-level optimizations. Our survey focuses on the three different level resource-efficient techniques for CNN architectures, since CNN is one of the most widely used deep learning architectures \cite{krizhevsky-imagenet}. 

Related survey works are as follows.
Sze et al. \cite{sze-efficient} provided a comprehensive tutorial and survey towards efficient processing of DNNs, discussing DNN architectures, software frameworks (e.g., PyTorch, TensorFlow, Keras, etc.), and the implementation methods optimizing Multiply-and-Accumulate Computations (MACs) of DNNs on given compute platforms.
Cheng et al. \cite{cheng-model} conducted a survey on the model compression techniques including pruning, low-rank factorization, compact convolution, and knowledge distillation. Deng et al.~\cite{deng-model_compression} discussed joint model-compression methods which combined multiple model-level compression techniques, and their efficient implementation on particular computing platforms. Wang et al. \cite{wang-deep} provided a survey on custom hardware implementations of DNNs and evaluated their performances using the Roofline model of \cite{williams-roofline}.

Unlike the previous survey works, we conduct a comprehensive survey on resource-efficient deep learning techniques in terms of the model-, arithmetic-, and implementation-level techniques by clarifying which resource-efficiency can be improved with particular techniques according to our resource-efficiency metrics as defined in Section~\ref{subsec:resource-eff-metric}. Such clarification would provide machine learning engineers, computer arithmetic designers, software developers, and hardware manufacturers with useful information to improve particular resource efficiency for their DNN applications. 
Besides, since we notice that fast wireless communication and edge computing development affects deep learning applications \cite{zhang-deep}, our survey also includes cutting-edge resource-efficient techniques for distributed AI such as early exiting techniques \cite{teer-branchy, teer-distributed}. The holistic and multi-facet view for resource-efficient techniques for deep learning from our survey would allow for a better understanding of the available techniques and, as consequence, a better global optimization, compared to previous survey works. 
The main contributions of our paper include:  

\begin{itemize}
\item This paper first provides a comprehensive survey coverage of the recent resource-efficient techniques for DNNs {in terms of the model-, arithmetic-, and implementation-level techniques}.
\item To the best of our knowledge, our paper is the first to provide comprehensive survey on arithmetic-level utilization techniques for deep learning. 
\item This paper utilizes multiple resource efficiency metrics to clarify which resource efficiency metrics each technique improves. 
\item This paper provides the influence of resource-efficient deep learning techniques from higher to lower level techniques (refer to Fig.~\ref{fig:survey-diagram}).  
\item We discuss the future trend for the resource-efficient deep learning techniques. 
\end{itemize} 

We discuss our resource efficiency metrics for deep learning in Section~\ref{sec:res-efficiency}, the model-level resource-efficient techniques in Section~\ref{sec:model-level}, the arithmetic-level techniques in Section~\ref{sec:arith-level}, the implementation-level techniques in Section~\ref{sec:implementation}, the influences between different-level techniques in Section~\ref{sec:influence}, the future trend in Section~\ref{sec:open issues}, and conclusion in Section~\ref{sec:conclusions}. Our paper excludes higher-level training procedure manipulation techniques such as one-pass ImageNet \cite{hu-one-pass}, bag of freebies \cite{bochkovskiy2020yolov4}, data augmentation, etc.

\section{Background on Deep Learning and Resource-Efficiency}    \label{sec:res-efficiency}
This section describes deep learning overview and resource efficiency metric, as preparatory to the description of resource-efficient techniques via the three different levels. 

\subsection{Deep Learning Overview}
Deep learning is defined as ``learning multiple levels of representation'' \cite{bengio-deep_learning} and often utilizes DNNs to learn the multiple levels of representation. DNNs are trained using the training data set, and their prediction accuracy is evaluated using the test dataset \cite{abu}. In this section, we describe the perceptron model (i.e., artificial neuron) first and then DNNs later.  

\subsubsection{Perceptron Model:}
\begin{figure}[!t] 
\centering
\includegraphics[width=4.8in]{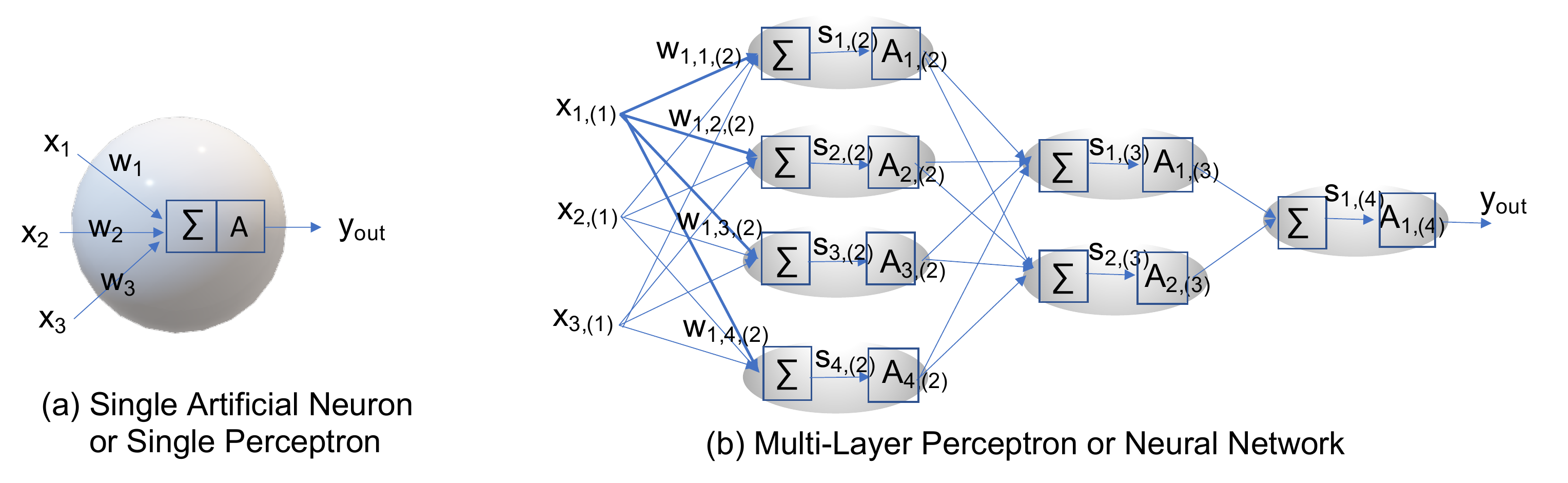}
\caption{Perceptron and neural network model.}
\label{fig:perceptron}
\end{figure}

The McCulloch and Pitts's neuron (a.k.a. M-P neuron) \cite{mcculloch-logical}, proposed in 1943, was a system mimicking the neuron in the nervous system, receiving multiple binary inputs and producing one binary output based on a threshold. Inspired by the work of \cite{mcculloch-logical}, Rosenblatt \cite{rosenblatt-perceptron} proposed the ``perceptron'' model consisting of multiple weights, a summation, and an activation function as shown in Fig.~\ref{fig:perceptron}.(a).
Eq.~(\ref{eq:perceptron}) describes a perceptron's firing activity $y_{out}$ using the inputs $x_i$ associated with their weights $w_i$, where the $i$ represents an index to indicate one of multiple inputs.
\begin{equation} \label{eq:perceptron}
y_{out} = 
\begin{cases} 
      1, & \text{if } (\Sigma_{i=1}^{n_{in}} w_i \times x_i > threshold) \text{ or } (\Sigma_{i=1}^{n_{in}} w_i \times x_i + bias > 0) \\
      0, & \text{if } (\Sigma_{i=1}^{n_{in}} w_i \times x_i \leq threshold) \text{ or } (\Sigma_{i=1}^{n_{in}} w_i \times x_i + bias \leq 0),
\end{cases}
\end{equation}
where $n_{in}$ is the number of the inputs. 
The function that determines the firing activity is referred to as the \textit{activation function}, and the \textit{bias} is in proportion to the probability of the firing activation \cite{nielsen-neural}. Since single perceptron model is suitable only for linearly separable problems, a Multi-Layer Perceptron (MLP) model can be used for non-linearly separable problems as shown in Fig.~\ref{fig:perceptron}.(b), where $w_{j,k,(i)}$ represents a weight linking the $j$\textsuperscript{th} neuron in the $(i-1)$\textsuperscript{th} layer to the $k$\textsuperscript{th} neuron in the $i$\textsuperscript{th} layer. 
The signal $s_{j,(i)}$ in Fig.~\ref{fig:perceptron} follows Eq.~(\ref{eq:signal}):
\begin{equation}
\label{eq:signal}
s_{j,(l)} = \Sigma_{i=1}^{n_{in}^{(l-1)}} (w_{i,j,(l)} \times x_{i,(l-1)}) = (\mathbf{W}_{(l)}^T \mathbf{x}_{(l-1)})_j,   
\end{equation}
and $x_{j,(l)} = \theta_{P}(s_{j,(l)})$, where $\theta_{P}(s)$ is a perceptron's activation function that follows Eq.~(\ref{eq:perceptron}) (i.e., step function), and $\mathbf{W}_{(l)}$ consists of the matrix elements, $w_{i,j,(l)}$s, for the $i^{th}$ row and the $j^{th}$ column.   
\subsubsection{Deep Neural Network:}
Since it requires tremendous efforts for human to optimize MLPs manually, neural networks that adopts a soft threshold \textit{activation function} $\theta_N$ (e.g., $sigmoid$, $ReLU$, etc.) were proposed to train the weights according to the training data \cite{wilson-excitory, werbos-backpropagation}.
\cite{nielsen-neural} notice that neural network is sometimes interchangeably used with MLP. For clarity, we name an algorithm as an MLP if it utilizes a step function for its activation functions and as a neural network if it utilizes a soft threshold function. In Fig.~\ref{fig:perceptron}.(b), the output from the $i$\textsuperscript{th} neuron at the $l$\textsuperscript{th} layer in a neural network employing a soft threshold activation function, $\theta_N(\cdot)$,  can be represented as Eq.~(\ref{eq:forward_pass}):  
\begin{equation}
\label{eq:forward_pass}
x_{i,(l)} = \theta_N(s_{i,(l)}).   
\end{equation}
A neural network allows the weights and the biases to be trained using the backpropagation \cite{abu}.
A neural network model is often referred to as a \textit{feed-forward} model in that the weights always link the neurons in the current layer to the neurons in the very next layer.
In a neural network, the middle layers located between the input and output layer, are often referred to as \textit{hidden layers} (e.g., two hidden layers in Fig.~\ref{fig:perceptron}.(b)). A neural network with multiple hidden layers is referred to as a \textit{DNN} \cite{sze-efficient}. 
   
\subsubsection{Training - Backpropagation:}
In the forward pass, the neuron outputs are propagated in forward direction based on the matrix-vector multiplications as shown in Eq.~(\ref{eq:signal}). Likewise, the weights and the biases can be trained in backward direction using matrix-vector multiplications. This method is called as the backpropagation. The backpropagation method consists of the three steps, allowing a gradient descent algorithm to be implemented efficiently on computers. It finds the activation gradients, $\delta_{j,(l)}$s (i.e., the gradients with respect to all the signals, $s_{j,(l)}$s, in Eq~(\ref{eq:signal})), in step 1, finds the weight gradients (i.e., the gradients with respect to all the weights)  using the activation gradients in step 2, and finally updates the weights using the weight gradients in step 3. All $\delta_{j,(l-1)}$s are found in backward direction using the matrix-vector multiplications by replacing $\mathbf{W}_{(l)}^T$ to $\mathbf{W}_{(l)}$ and $x_{j,(l-1)}$ to $\delta_{j,(l)}$ in  Eq.~(\ref{eq:forward_pass}). After all activation gradients have been found, the weight gradients can be found. Finally, the weights are updated using the weight gradients. The backpropagation requires additional storage to store all the weights and activation values. Once a DNN is trained, the DNN is used for the inference task using the trained weights. Please refer to \cite{abu} for the further details for the backprogation method. After a DNN being trained, the DNN's accuracy is evaluated using the validation dataset which is unseen from the training.

\subsubsection{Convolutional Neural Network:}
\begin{figure}[!t] 
\centering
\includegraphics[width=3.8in]{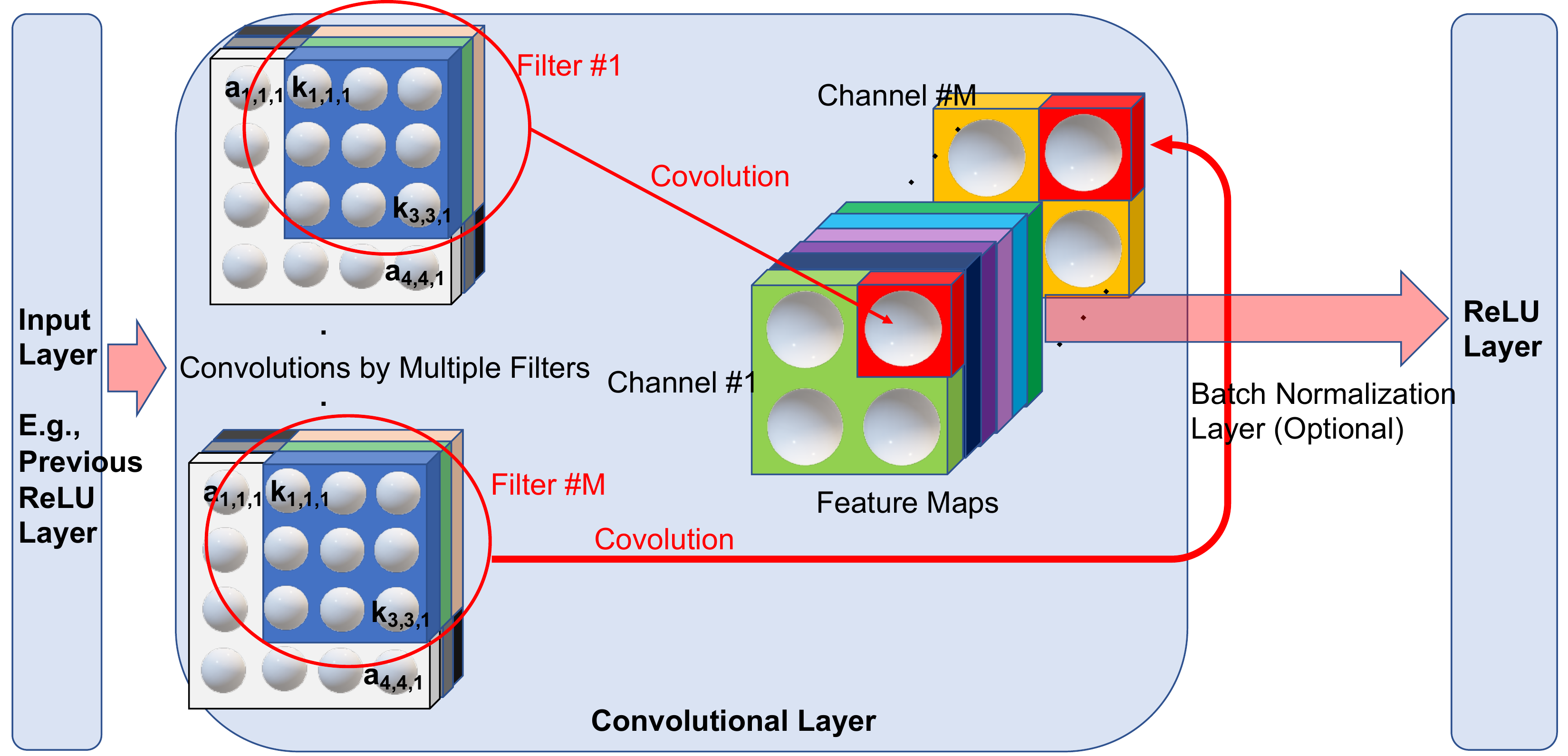}
\caption{Convolution operations in a CNN.}
\label{fig:convolution}
\end{figure}

Since CNN is one of the most successful and widely used deep learning architectures \cite{krizhevsky-imagenet}, we exemplify CNN as a representative deep learning architecture. CNN employs multiple convolutional layers, and each convolutional layer utilizes multiple filters to perform convolutions independently with respect to each filter as shown in  Fig.~\ref{fig:convolution}, where a \textit{filter} at a convolutional layer consists of as many \textit{kernels} as the number of the channels at the input layer (e.g., 3 kernels per filter in Fig.~\ref{fig:convolution}).
For example, each $3 \times 3$ filter has 9 weight parameters and slides from the top-left to the bottom-right position, generating $4$ output values with respect to each position (e.g., top-left, top-right, bottom-left, and bottom-right position) in Fig.~\ref{fig:convolution}. The outputs from the convolutions are often called \textit{feature maps} and are fed into activation functions. Modern CNNs such as ResNet \cite{he-deep} often employ a batch normalization layer \cite{ioffe-batch-normalization} between the convolutional layer and the ReLU layer to improve the accuracy.

The CNN is a resource-efficient deep learning architecture in terms of the accuracy per parameter and per operation by leveraging the two properties, \textit{local receptive field} and \textit{shared weights} \cite{lecun-backpropagation}. For example, performing convolutions using multiple small kernels extracts multiple local receptive features from the input image during training, and each kernel contains some meaningful pattern from the input image after being trained \cite{zeiler-visualizing}. Thus, CNN utilizes much fewer weights than fully connected DNN, since the kernel's height and weight are generally much smaller than the height and the width at the input layer, leading to the improved resource efficiency. Notice that a convolutional layer becomes a fully connected layer if the height and the width at the input layer are matched with each kernel's height and width. The number of total weights in a layer in a CNN is much less than used in a fully connected neural network, since the local receptive weights are shared over the entire feature on a layer. 

Training CNN also utilizes the backprogation using the transpose of kernel matrices in a filter to update the weights in the filter. The mini-batch gradient descent algorithm is widely used to train CNNs, which utilizes part of training data to update the weights per iteration. The number of data used per iteration is often referred to as the \textit{batch size} $B$ (e.g., $B = 64$ or $128$). Each $epoch$ consumes the entire training data, consisting of $N/B$ iterations, where $N$ is the number of the entire training data.   
The mini-batch gradient descent method is a resource-efficient training algorithm in terms of the accuracy per operation, compared to the batch gradient descent method that utilizes entire training dataset per iteration (i.e., the batch gradient descent method updates the weights per epoch). 
For parallel backpropagation implementation with respect to $B$ data samples in one mini-batch, all the weights and all the activation values using $B$ training samples should be stored to update the weights per the mini-batch iteration, requiring $B \times$ additional storage, compared to the backpropagation using a stochastic gradient descent algorithm which updates the weights per training sample. Our paper refers to the term, \textit{DNN}, as any neural network with several hidden layers, including CNN.  

\subsection{Resource Efficiency Metrics for Deep Learning} \label{subsec:resource-eff-metric}
Recently, researchers from DeepMind \cite{hu-one-pass} proposed the metrics for resource-efficient deep learning benchmarks, including the top-1 accuracy, the required memory footprint for training, and the required number of floating operations for training, and evaluated the resource-efficiency for deep learning applications with jointly considering the three metrics. The Roofline model \cite{williams-roofline} discussed attainable performance in terms of the operational intensity defined as the number of floating point operations per DRAM access. Motivated by \cite{hu-one-pass, williams-roofline}, our resource efficiency metrics include the accuracy per parameter, per operation, per memory footprint, per core utilization, per memory access, and per Joule as shown in Fig.~\ref{fig:survey-diagram}. 

\subsubsection{Accuracy per Parameter:}
We consider the accuracy per parameter (i.e., weight) for a resource-efficiency metric. The accuracy per parameter is an abstract resource efficiency metric since higher accuracy per parameter does not always imply higher physical resource efficiency after its implementation \cite{hu-one-pass, imagenet-benchmark}.

\subsubsection{Accuracy per Operation:} We consider the accuracy per arithmetic operation for a resource-efficiency metric. This is also an abstract metric, since it can be evaluated prior to the implementation.

\subsubsection{Accuracy per Compute Resource:}
The instruction-driven architecture such as CPU or GPU requires substantial memory accesses due to instruction fetch and decode operations, while the data-driven architecture such as ASIC or FPGA can minimize the number of memory accesses, resulting in energy efficiency. We further categorize such compute resource into core utilization, memory footprint, and memory access, required to operate a DNN on given computing platforms. For example, the memory access can be interpreted as GPU DRAM access (or off-chip memory) for a GPU and as FPGA on-chip memory access (or off-chip memory) for a FPGA.

\textit{a. Accuracy per Core Utilization:} The core utilization in this paper represents the utilization percentage of the processing cores or processing elements.

\textit{b. Accuracy per Memory Footprint:}
The accuracy per memory footprint is related to both physical and abstract resource efficiency as shown in Fig.~\ref{fig:survey-diagram}. The memory footprint is in proportion to the number of the parameters, but it can be varied according to a precision-level applied for arithmetic. For example, if a half precision arithmetic is applied for a deep learning, the memory footprint can be saved by $2\times$, compared to a single precision arithmetic deep learning. 

\textit{c. Accuracy per Memory Access:} 
A computing kernel having a low operational intensity cannot approach a peak performance defined by hardware specification since the data supply rate from DRAM to CPU cannot catch up with the data consumption rate by arithmetic operations. Such kernels are called ``memory bound kernels'' in \cite{williams-roofline}. Other type kernels are named ``compute bound kernels'' that can approach a peak performance defined by hardware specification. Utilizing reduced precision arithmetic can improve the performance for both memory bound kernels by improving the data supply rate from DRAM to CPU and compute bound kernels by increasing word-level parallelism on SIMD architectures \cite{lee-air}. 

\subsubsection{Accuracy per Joule:} \label{subsubsec:A/J}
The dynamic power consumption is the main factor to determine energy consumption required for computationally intensive tasks (e.g., DNN training/inference tasks). 
The dynamic power consumption, $P_D$, follows:
\begin{equation}
    P_D = \#_{TTR} \times C_{CP} \times V^{2}_{CP} \times f_{CP}, 
\end{equation}
where $\#_{TTR}$ is the number of toggled transistors, $C_{CP}$ is an effective capacitance, $V_{CP}$ is an operational voltage, and $f_{CP}$ is an operational frequency for a given computing platform $CP$. Generally, the required minimum operational voltage is in proportion to the operational frequency. Therefore, adapting the frequency to the voltage scaling can save power cubically (a.k.a. Dynamic Voltage Frequency Scaling \cite{shang-dynamicpower}). For example, minimizing the operations required to operate DNN during runtime contributes to minimizing $\#_{TTR}$, resulting in power reduction and energy saving; we discuss further the resource-efficient techniques leveraging this in Section \ref{sec:skipping operations}.   

\section{Model-Level Resource-Efficient Techniques} \label{sec:model-level}
\begin{figure}[!t] 
\centering
\includegraphics[width=\linewidth]{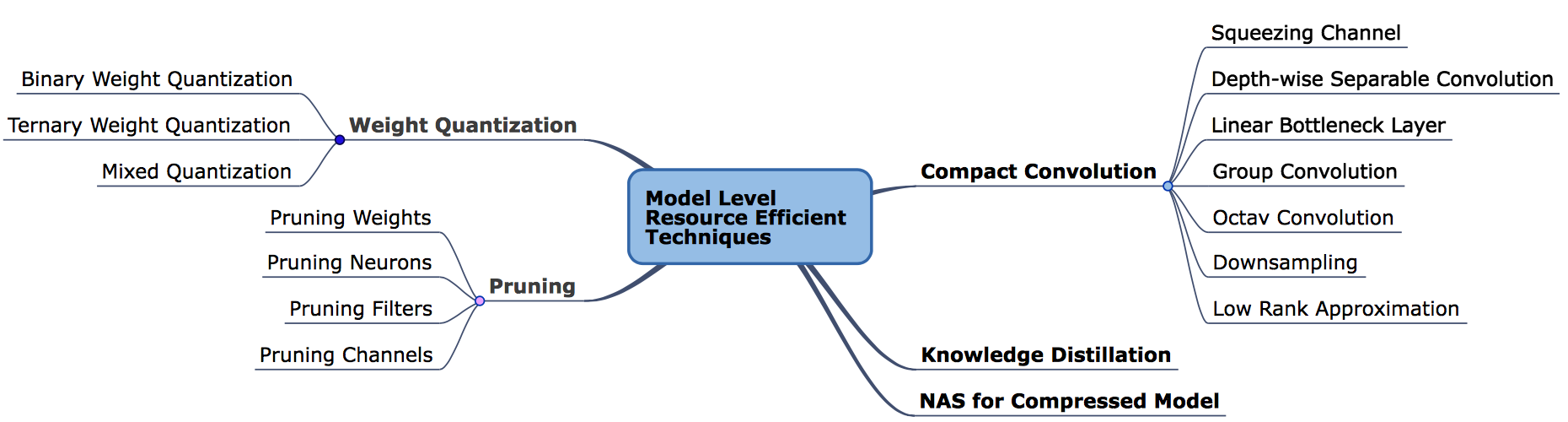}
\caption{Categorization for model-level resource-efficient techniques.}
\label{fig:model_level_diagram}
\end{figure}

The model-level resource-efficient techniques, mostly developed from machine learning community, aim at reducing the DNN model size to fit the models to resource-constrained systems such as mobile devices, IoTs, etc. 
We categorize the model-level resource-efficient techniques as shown in Fig.~\ref{fig:model_level_diagram}. 

\subsection{Weight Quantization}
The weight quantization techniques quantize the weights with a smaller number of bits, improving the accuracy per memory footprint. The training procedure should be amended according to the weight quantization schemes.

\subsubsection{Binary Weight Quantization:} 
BinaryConnect training scheme~\cite{courbariaux-binnaryconnect} allowed a DNN to represent the weights using one bit. In step 1, the weights are encoded to $\{-1, 1\}$ using a stochastic clipping function. In step 2, the forward pass is performed using the encoded binary weights. In step 3, backpropagation seeks all activation gradients using full precision. In step 4, the weights are updated using full precision, and the training procedure goes back to step 1 for the training using the next mini-batch. This method required only one bit to represent the weights, thus improving the accuracy per memory footprint. In addition, the binary weight quantization also removed the need for multiplication arithmetic operations for MAC operations, improving the accuracy per operation. Moreover, if the activations are also quantized to the binary value, all MAC operations in DNN can be implemented only with XNOR gates and a counter \cite{courbariaux-binarized, redfern2021bcnn}.    

\subsubsection{Ternary Weight Quantization:}
Li et al. \cite{li-ternary} proposed ternary weight networks that utilized ternary weights, improving accuracy, compared to the binary weight networks.  All the weights on each layer were quantized into three values, requiring only two bits to represent the quantized weights. Overall training procedure was similar to \cite{courbariaux-binnaryconnect}, but with ternary valued weights instead of the binary weights. 
The ternary weight network showed equivalent accuracy to various single precision networks with MNIST, CIFAR-10 and ImageNet, while the binary weight quantization \cite{courbariaux-binnaryconnect} showed minor accuracy loss. 
Zhu et al. \cite{zhu-ternary} scaled the ternary weights independently for each layer with a layer-wise scaling approach, improving the accuracy further, compared to \cite{li-ternary}. 

\subsubsection{Mixed Quantization:}
\cite{hubara-quantized} proposed ``Quantized Neural Network'' that quantizes the activations and the weights to arbitrary lower precision format. For example, quantizing the weights to 1 bit and the activations to 2 bits improved the accuracy, compared to the binarized DNN of \cite{courbariaux-binarized}. 

\subsection{Pruning}
Pruning unimportant neurons, filters, and channels can save computational resources for deep learning applications without sacrificing accuracy, improving the accuracy per parameter and per operation. Coarse-grained pruning methods such as pruning filters or channels are not flexible to achieve a prescribed accuracy, but can be implemented efficiently on hardware \cite{liu-learning}, implying higher physical resource efficiency than fine-grained pruning such as pruning weights. Notice that such pruning methods can degrade confidence scores without careful re-training, even though they did not affect top-1 accuracy \cite{yazdani-dark}. 

\subsubsection{Pruning Weights:}
In 1990, LeCun et al. proposed a weight pruning method to generate sparse DNNs with fewer weights without losing accuracy \cite{lecun-optimal}. In 2015, the weight pruning approach was revisited \cite{han-learning}, and the weights were pruned based on their magnitudes after training – the pruned DNNs were retrained to regain the lost accuracy. The pruning and re-training procedures could be performed iteratively to prune the weights further. This method reduced the number of weights of AlexNet by $9 \times$ without losing accuracy. In 2016, Guo et al.~\cite{guo-dynamic} noticed that pruning wrong weights could not be revived, and proposed to prune and splice the weights per mini-batch training to minimize the risk from pruning wrong weights from previous mini-batch training. For example, the pruned weights were also participated in the weight update procedure during the backpropagation and were restored when they were re-considered as the important weights. In 2017, Yang et. al \cite{Yang-designing} proposed an energy-aware weight pruning method in which the energy consumption of a CNN was directly measured to guide the pruning process. In 2019, Frankly et al. \cite{frankle-lottery} demonstrated that some of pruned models outperformed the original model by retraining the pruned models with replacing the survived weights with the initial random weights used for training the original model. 

\subsubsection{Pruning Neurons:}    
Instead of pruning individual weights, pruning a neuron can remove a group of the weights belonging to the neuron \cite{srinivas-data-free, mariet-diversity, hu-network, yu-nisp}. In 2015, \cite{srinivas-data-free} pruned the redundant neurons having similar weight values in a trained DNN model. For example, the weights in a baseline neuron were compared to the weights in other neurons at the same layer, and the neurons having similar weights to the baseline neuron were fused to the baseline neuron based on a Euclidean distance metric in the weight values between the two neurons. In 2016, \cite{mariet-diversity} pruned the redundant neurons based on the ``determinantal point process'' metric. Hu et. al~\cite{hu-network} measured the average percentage of zero activations per neuron and pruned the neurons having a high percentage of zero activations according to a given compression rate. Yu et al. \cite{yu-nisp} pruned unimportant neurons based on the effect of the pruning error propagation on the final response layer (e.g., the neurons were pruned backward from the final layer to the first layer). The methods of pruning neurons improved the resource efficiency such as the accuracy per parameter and per operation. 

\subsubsection{Pruning Filters:} 
Pruning insignificant filters after training can improve the accuracy per parameter and per operation. The feature maps associated with the pruned filters and the next kernels associated with the pruned feature maps should be also pruned. Pruning filters can maintain the dense structure of DNN unlike pruning weights, implying that it is highly probable to improve physical resource efficiency further, compared to pruning weights. Li et al. \cite{li-pruning} pruned unimportant filters based on the summation of absolute weight values in the filter. 
The pruned DNNs were retrained with the survived filters to regain the lost accuracy. 
Yang et al. \cite{yang-netadapt} pruned filters based on a platform-aware magnitude-based metric depending on the resource-constrained devices. 
\textit{ThiNet} \cite{luo-thinet} calculated the significance of the filters using the outputs of the next layer and pruned the insignificant filters based on this significance measurement. 

\subsubsection{Pruning Channels:} 
Unlike pruning filters, pruning channels removes the filters at the current layer and the kernels at the next layer associated with the pruned channels. The network slimming approach \cite{liu-learning} pruned insignificant channels, producing compact models while keeping equivalent accuracy, compared to the models prior to pruning. For example, insignificant channels were identified based on scaling factors generated from the batch normalization of \cite{ioffe-batch-normalization}, and the channels associated with lower scaling factors were pruned. After the initial training, the channels associated with relatively low scaling factors were first pruned, and retraining was then performed to refine the network. 
He et al. \cite{he-channel} identified unimportant channels using LASSO regression from a pre-trained CNN model and pruned them. The channel pruning brought 5 $\times$ speed-up on VGG-16 with minor accuracy loss. 
Lin et al. \cite{lin-runtime} pruned unimportant channels during runtime based on a decision maker trained by reinforcement learning.
Gao et al. \cite{gao-dynamic} proposed another dynamic channel pruning method that dynamically skipped the convolution operations associated with unimportant channels.

\subsection{Compact Convolution}
To improve resource-efficiency such as the accuracy per operation and per parameter from computationally intensive convolution operations, many compact convolution methods were proposed. 

\subsubsection{Squeezing Channel:}
In 2016, Iandola et al. \cite{iandola-squeezenet} proposed \textit{SqueezeNet} in which each network block utilized the number of 1$\times$1 filters less than the number of the input channels to reduce the network width in the squeezing stage and then utilized multiple 1$\times$1 and 3$\times$3 kernels in the expansion stage. 
The computational complexity was significantly reduced by squeezing the width, while compensating the accuracy in the expansion stage. SqueezeNet reduced the number of parameters by $50 \times$, compared to AlexNet on ImageNet without losing accuracy, improving accuracy per parameter.  
Gholami et al. \cite{gholami-squeezenext} proposed \textit{SqueezeNext} that utilized separable convolutions in the expansion stage; a $k \times k$ filter was divided into a $k \times 1$ and a $1 \times k$ filter. Such separable convolutions reduced the number of parameters further, compared to SqueezeNet while maintaining AlexNet's accuracy on ImageNet, improving accuracy per parameter further, compared to SqueezeNet.   

\subsubsection{Depth-Wise Separable Convolution:}
\begin{figure}[!t] 
\centering
\includegraphics[width=3.8in]{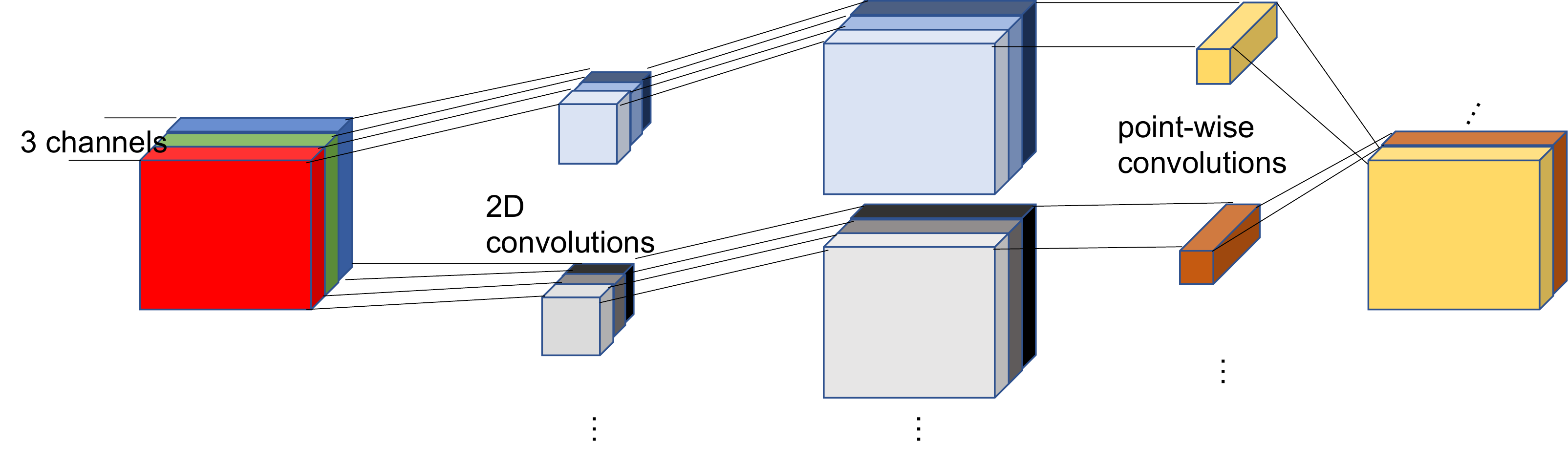}
\caption{Depth-wise convolution used in \cite{howard-mobilenets}.}
\label{fig:dwconv}
\end{figure}

\textit{Xception} \cite{Chollet-xception} utilized the depth-wise separable convolutions, that replace 3D convolutions with 2D separable convolutions followed by 1D convolutions (i.e., point-wise convolutions) as shown in Fig.~\ref{fig:dwconv}, to reduce computational complexity. The 2D separable convolutions are performed separately with respect to different channels. Howard et al. \cite{howard-mobilenets} proposed \textit{MobileNet v1} that utilizes the depth-wise separable convolutions with the two hyperparameters, ``width multiplier and resolution multiplier",  to fit  DNNs to resource-constrained devices by fully leveraging the accuracy and resource trade-off in the DNNs. MobileNet v1 showed equivalent accuracy to GoogleNet and VGG16 on ImageNet dataset with less computational complexity, improving the accuracy per parameter and per operation. 

\subsubsection{Linear Bottleneck Layer:} 
\begin{figure}[!t] 
\centering
\includegraphics[width=3.3in]{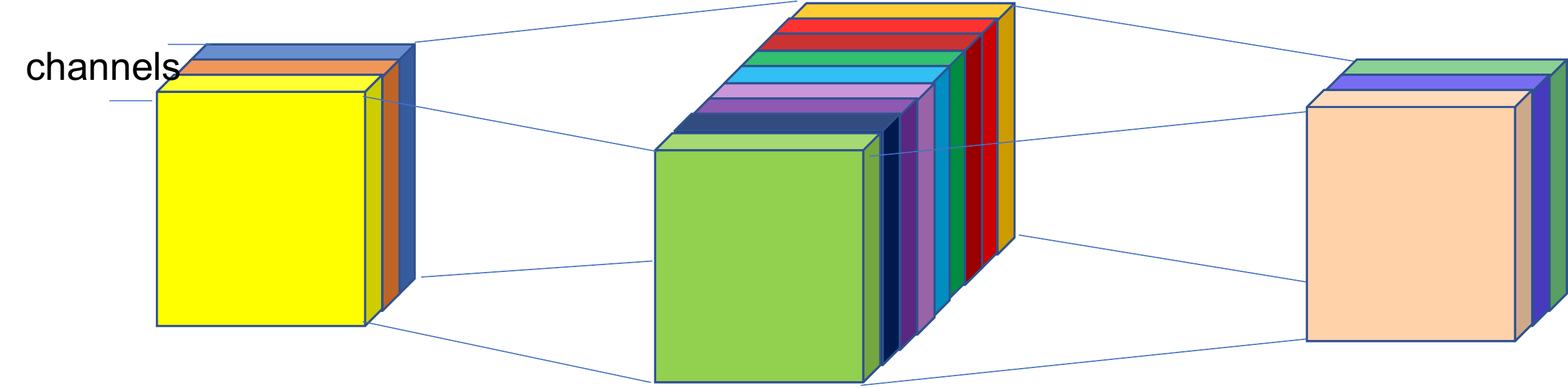}
\caption{Bottleneck layer block used in \cite{sandler-mobilenetv2}.}
\label{fig:bottleneck}
\end{figure}

In general, the manifold of interest (i.e., the subspace formed by the set of activations at each layer) could be embedded in low-dimensional subspaces in deep learning. 
Inspired by this, Sandler et al. \cite{sandler-mobilenetv2} proposed MobileNet v2 consisting of a series of bottleneck layer blocks. Each bottleneck layer block as shown in Fig.~\ref{fig:bottleneck} received lower dimensional input, expanded the input to high dimensional intermediate feature maps, and projected the high dimensional intermediate features onto low dimensional features. Keeping linearity for the output feature maps was crucial to avoid destroying information from non-linear activations, so linear activation functions were used at the end of each bottleneck block.
 
\subsubsection{Group Convolution:}
In a group convolution method, the input channels are divided into several groups, and the channels in each group are separately participated in convolution with other groups. For example, the input channels with three groups required three separate convolutions. Since group convolution does not communicate with the channels in other groups, communication between different groups is performed after the separate convolutions. Group convolution methods of \cite{zhang-shufflenet, ma-shufflenetv2, huang-condensnet, huang-densely} reduced the number of MAC operations, improving the accuracy per operation, compared to DNNs using regular convolution.  
In 2012, AlexNet utilized group convolution to train the DNNs effectively using the two NVIDIA GTX580 GPUs \cite{krizhevsky-imagenet}. Surprisingly, an AlexNet using the group convolution showed superior accuracy to an AlexNet using regular convolution, improving the accuracy per operation. In 2017, ResNext~\cite{xie-resnext} utilized group convolution based on ResNet~\cite{he-deep} using a cardinality parameter (i.e., the number of groups). In 2018, Zhang et al. \cite{zhang-shufflenet} noticed that the point-wise convolutions were computationally intensive in practice in the depth-wise convolutions and proposed {ShuffleNet} that applied group convolution to every point-wise convolution to reduce compute complexity further, compared to MobileNet v1. ShuffleNet shuffled the output channels from the grouped point-wise convolution to communicate with different grouped convolutions, demonstrating superior accuracy to MobileNet v1 on ImageNet and COCO datasets, given the same arithmetic operation cost budget. Ma et al. \cite{ma-shufflenetv2} proposed ShuffleNet v2 that improved physical resource efficiency further, compared to ShuffleNet \cite{zhang-shufflenet} by employing equal channel width for input and output channels where applicable and minimizing the number of operations required for $1 \times 1$ convolutions.  
Rather than choosing each group randomly and shuffling them, Huang et al. \cite{huang-condensnet} proposed to learn each group for a group convolution during training. The ``learned group convolution'' was applied in \textit{Densenet} \cite{huang-densely}, and Densenet improved the accuracy per parameter and per operation, compared to ShuffleNet, given a prescribed accuracy.

\subsubsection{Octave Convolution:}
Chen et al. \cite{chen-drop} decomposed feature maps into a higher and a lower frequency part to save the feature maps' memory footprint and reduce the computational cost. The decomposed feature maps were used by specific convolution called ``Octave Convolution'' that performs a convolution between the higher and lower frequency part. The application of the octave convolution to ResNet-152 architecture achieved higher accuracy using ImageNet dataset than the regular convolution, improving the accuracy per operation and per memory footprint.

\subsubsection{Downsampling:} Qin et al. \cite{qin-fd-mobilenet} applied a downsampling approach (e.g., a larger stride size for a convolution) to MobileNet v1, improving the top-1 accuracy by 5.5\% over MobileNet v1 on the ILSVRC 2012 dataset, given a 12M arithmetic operations budget.

\subsubsection{Low Rank Approximation:}
Denton et al. \cite{denton-exploiting} proposed a low rank approximation that compresses the kernel tensors in the convolutional layers and the weight matrices in the fully connected layers by using singular value decomposition. Another low rank approximation \cite{kim-compression} used Tucker decomposition to compress the feature maps, resulting in significant reductions in the model size, the inference latency, and the energy consumption. Such low rank approximation methods improve the accuracy per parameter, per operation, and per memory footprint.

\subsection{Knowledge Distillation} 
The knowledge from a large-scale high performing model (teacher network) could be transferred to a compact neural network (student network) to improve resource efficiency such as accuracy per parameter and per operation for inference tasks \cite{bucilua-model, romero-fitnets, chen-learning}.  
Bucilu\u{a} et al. \cite{bucilua-model} utilized data with the labels generated from the teacher model (i.e., a large scale ensemble model) to train a compact neural network.
The compact model was trained with the pseudo training data generated from the teacher model, demonstrating equivalent accuracy to the teacher model. Ba and Caruana \cite{ba-do_deep} noticed that the softmax outputs often resulted in the student network ignoring the information of the other categorizations than the one with the highest probability, and utilized the values prior to the softmax layer, from the teacher network, for the training labels to allow the student network to learn the teacher network more efficiently. Hinton et al. \cite{hinton-distilling} added a ``temperature'' term for the labels to enrich the information from the teacher network and train the student network more efficiently, compared to \cite{ba-do_deep}.
Romeo et. al \cite{romero-fitnets} utilized both labels and intermediate representations from a wider teacher network to compress it to a thinner and deeper student network. The ``hint layer'' was chosen from the teacher network and the ``guided layer'' was chosen from the student network. The student network was then trained so that the intermediate representation deviation between the outputs from the hint layers and guided layers could be minimized. A thinner student network employed $10.4 \times$ less weight parameters, compared to a wider teacher network, while improving accuracy. This technique is also known as ``hint learning''.  The hint learning was applied to both the region proposal and classification components for object detection applications \cite{chen-learning}. 

\subsection{Neural Architecture Search for Compressed Models}
Zoph et al. \cite{zoph-neural} proposed Neural Architectural Search (NAS) technique to seek optimal DNN models in the space of hyperparameters of network width, depth, and resolution.
In case that compute resource budget was limited (e.g., mobile devices), many NAS variants exploited the trade-off between accuracy and latency to maximize resource efficiency given compute resource budget \cite{he-amc, Tan-mnasnet, tan-efficientnet, tan-efficientdet, wu-fbnet, anderson-performance}.         
He et al. \cite{he-amc} proposed a NAS employing reinforcement learning, \textit{AutoML}, 
that sampled the least sufficient candidate design space to compress the DNN models. \textit{MnasNet}~\cite{Tan-mnasnet} utilized reinforcement learning with a balanced reward function between the accuracy and the latency to seek a compact neural network model. 
Wu et. al \cite{wu-fbnet} proposed a gradient-based NAS that produced a DNN model with $2.4\times$ model size reduction, compared to a MobileNet v2 without losing accuracy on ImageNet dataset. Florian et. al \cite{florian-constrained} proposed a narrow-space NAS 
to generate low-resource DNNs satisfying strict memory budget and inference time requirement for IoT applications. 
\cite{anderson-performance} noticed that conventional NAS might improve abstract resource efficiency rather than physical resource-efficiency, and utilized the hardware information including the inference latency for a NAS to ensure that the candidate models could improve the physical resource-efficiency in practice. \textit{Efficientnet} \cite{tan-efficientnet} utilized a NAS with compound scaling of depth, width, and resolution to seek optimal DNN models given fixed compute resource budgets. Another NAS utilizing compound scaling, \textit{Efficientdet}, was proposed for object detection applications \cite{tan-efficientdet}.
Efficientdet improved the accuracy using COCO dataset with $4-9\times$ model size reduction, compared to state-of-the-art object detectors, improving the accuracy per parameters. 
Recently, \cite{cai-ofa} proposed a feed-forward NAS approach that produced a customized DNN, given compute resource and latency constraint. 

\section{Arithmetic-Level resource-efficient Techniques} \label{sec:arith-level}
\begin{figure}[!t] 
\centering
\includegraphics[width=\linewidth]{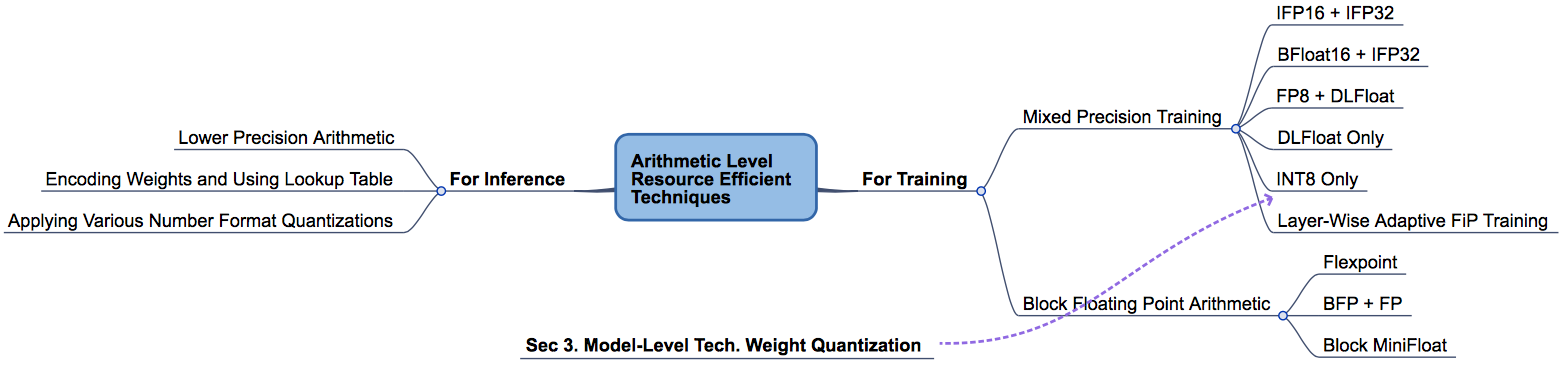}
\caption{Categorization of arithmetic-level resource-efficient techniques.}
\label{fig:arithmetic_level_diagram}
\end{figure}

Utilizing lower precision arithmetic reduces the memory footprint and the time spent transferring data across buses and interconnections \cite{fox-blockminifloat, micikevicius-mixed, zhu-towards-int8, YANG-training}. Employing least sufficient arithmetic precision for DNN applications can improve the accuracy per memory footprint and the accuracy per memory access. We categorize the arithmetic-level resource-efficient techniques into the two categories as shown in Fig.~\ref{fig:arithmetic_level_diagram}, \textit{Arithmetic-Level Techniques for Inference} and \textit{Arithmetic-Level Techniques for Training}. We discuss different number formats first and the deployment of such number formats on DNNs later. 

\subsection{Number Formats for Deep Learning}
This subsection describes various number formats for deep learning applications, as preparatory to explaining the arithmetic-level resource-efficient techniques. 
Fixed-Point (FiP) number format utilizes a binary fixed point between fraction and integer part. For example, an 8-bit FiP format, ``01.100000'',  represents 1.5 (i.e., $ ... 0\times2^{1}+ 1\times2^0 + 1\times2^{-1} + 0\times2^{-2} ...$) for the decimal representation, and the point between integer part and fraction part is fixed for arithmetic operations. Therefore, it could be implemented with simple circuits, but the available data range is very limited \cite{parhami}.

We exemplify the IEEE-754 general-purpose floating-point standard \cite{IEEE754} to explain Floating-Point (FP) format and its arithmetic, since this standard is used for most commercially available CPUs and GPUs.
The IEEE 754 Floating-Point (IFP) data format \cite{IEEE754} consists of sign, exponent, and significand as shown in Eq.~(\ref{eq:float_format}). For example, a floating point number has a $(p+1)$-bit significand (including the hidden one), an $e$-bit exponent, and an $1$ sign bit.  The machine epsilon $\epsilon_{mach}$ is defined as $2^{-(p+1)}$. The value represented by FP is as follows:
\begin{equation}
 \label{eq:float_format}
y_{out} = \begin{cases} 
      \text{normal mode:} &  (-1)^{sign}\times(1\times2^0 + d_1\times2^{-1}+ ... + d_p\times2^{-p}) \times 2^{exponent - bias}\\
      \text{subnormal mode:} & (-1)^{sign}\times(d_1\times2^{-1}+ ... + d_p\times2^{-p}) \times 2^{1 - bias},
   \end{cases}
\end{equation}
where $d_1$, ... , $d_p$ represent binary digits, the `$1$' associated with the coefficient $2^0$ is referred to as the hidden `$1$', the $exponent$ is stored in offset notation, and the $bias$ is a positive constant. If the absolute value of exponent is zero, the floating-point value is represented by the subnormal mode.  
IEEE 754 standard requires exact rounding for addition, subtraction, multiplication, and division; the floating point arithmetic result should be identical to the one obtained from the final rounding after exact calculation. For example, based on the IEEE 754 rounding to nearest mode standard, floating point arithmetic should follow Eq.~(\ref{eq:flp_arith}): 

\begin{equation} \label{eq:flp_arith}
fl(x_1 \odot x_2) = (x_1 \odot x_2) (1 + \epsilon_r),
\end{equation} 
where $|\epsilon_r| \leq \epsilon_{mach}$, $\odot$ is one of the four arithmetic operations, and $fl(\cdot)$ represents the result from the floating point arithmetic. Notice that \textit{quantization} quantizes data to lower precision, while \textit{arithmetic} is a rule applied to arithmetic operations between the two operands. For example, the quantization affects the values for the two operands, $x_1$ and $x_2$ in Eq.~(\ref{eq:flp_arith}), while arithmetic affects the rounding error, $\epsilon_r$.  

\subsubsection{Half, Single, and Double Precision:}
The IEEE Floating-Point 32- (IFP32 or single precision) and 64-bit (IFP64 or double precision) versions are available on most of off-the-shelf conventional processors. Besides, IEEE-754 standard includes a 16-bit FP format (IFP16 or half precision) \cite{IEEE754}. $p=52$, $e=11$, and $bias = 1023$ for IFP64, $p=23$, $e=8$, and $bias=127$ for IFP32, and $p=10$, $e=5$, and $bias=15$ for IFP16. IFP16 is currently supported in hardware on some of modern GPUs to accelerate DNN applications  \cite{8091072, nvidia:v100}. 

\subsubsection{Brain Float-Point Format using 16 Bits (BFloat16):} In 2018, a 16-bit Brain Floating-Point format \cite{ying2018image, burgess-bfloat16} was proposed that was tailored to deep learning applications. The  BFloat16 consists of an 8-bit exponent and a 7-bit significand, supporting a wider dynamic data range than IFP16. 
BFloat16 is currently supported in hardware in the Intel Cooper Lake Xeon processors, the NVIDIA A100 GPUs, and the Google TPUs. 

\subsubsection{DLFloat:} In the race of designing specific FP formats for DNNs, \cite{agrawal-dlfloat, wang-training} proposed another 16-bit precision format, DLFloat, consisting of 6-bit exponent and 9-bit significand to provide better balance between dynamic data range and precision than IFP16 and BFloat16 formats for some of deep learning applications.

\subsubsection{TensorFloat32 (TF32):} 
NVIDIA proposed a 19-bit data format, TF32, consisting of 1-bit sign, 8-bit exponent and 10-bit significand to accelerate deep learning applications on A100 GPUs with the same dynamic range support as IFP32  \cite{fasi-higham}. TF32 Tensor cores in an A100 truncates IFP32 operands to 19-bit TF32 format but accumulates them using IFP32 arithmetic for MAC operations. 

\subsection{Arithmetic-Level Techniques for Inference} 
This subsection discusses various resource-efficient arithmetic-level techniques based on the pre-trained DNNs for the inference tasks.  

\subsubsection{Lower Precision Arithmetic:}
Lower precision FiP arithmetic has been widely used to deploy DNNs on edge devices \cite{yang2019quantization}. \cite{wu2020integer, wang2020unsupervised} analyzed the effect of deploying various lower precision arithmetic on the DNN inference tasks in terms of accuracy and latency. The \textit{BitFusion} method accelerated DNN inference tasks by employing variable bit-width FiP formats dynamically depending on the different layers~\cite{sharma-bit}. Similarly, Tambe et al. \cite {tambe-algorithm-hardware} proposed \textit{AdaptiveFloat} that adjusted dynamic ranges of FP numbers depending on the different layers, resulting in higher energy efficiency than FiP-based methods, given the same accuracy requirement. 

\subsubsection{Encoding Weights and Using Lookup Table:}
\cite{Bordawekar-abali} leveraged the fact that the exponent values of most of the weights were located within a narrow range and encoded the frequent exponent values of the weights with fewer bits using Huffman coding scheme, improving the accuracy per memory footprint for natural language processing applications. A lookup table, located between the memory and FP arithmetic units, is used to convert the encoded exponent values into FP exponent values.  

\subsubsection{Applying Various Number Format Quantizations to DNNs:}
The Residue Number System (RNS) is a parallel and carry-limited number system that transforms a big natural number to several smaller residues. Therefore, RNS was often used to perform parallel and independent calculations on residues without carry-propagation. \cite{7330131} exploited such parallelism to accelerate DNN computation. In a RNS-based DNN, the weights of a pre-trained model were transformed to RNS presentation. Recently, RNS was used to replace costly multiplication operations with simple logical operations such as multiplexing and shifting, accelerating DNN applications \cite{samimi-res-dnn, liu2020efficient, salamat-rnsnet}. The Logarithmic Number System (LNS) applies the logarithm to the absolute values of the real numbers \cite{parhami}. The main advantage of LNS is in the capability of transforming multiplications into additions and divisions into subtractions. In 2018, \cite{vogel-efficient} utilized a 5-bit logarithmic format using arbitrary log bases to improve the resource efficiency such as accuracy per memory footprint and per operations by replacing costly multiplication arithmetic operations to simple bit-shift operations \cite{vogel-efficient}.
The Posit number format \cite{gustafson-beating} employs multiple separate exponent fields to represent dynamic range effectively. Recently, DNNs utilizing Posit showed higher accuracy than various FP8 formats using Mushroom and Iris datasets \cite{carmichael-deepposition, carmichael-performance-efficiency}. 

\subsection{Arithmetic-Level Techniques for Training} 
This subsection discusses arithmetic-level resource-efficient techniques used for DNN training tasks. Training DNNs generally requires a higher precision arithmetic due to extremely small weight gradient values \cite{courbariaux-binnaryconnect, zhu-ternary}. Adjusting arithmetic precision according to different training procedures such as forward propagation, activation gradient updates, and weight updates can accelerate DNN training \cite{gupta-comparative}. Training quantized DNNs often required stochastic rounding schemes \cite{gupta, wu2018training, YANG-training, zhou-dorefanet}. 

\subsubsection{Mixed-Precision Training:} \label{sec:mixed-prec-training}
A conventional mixed precision training applied lower precision arithmetic to the multiplications in MACs, including both forward and backward path, and higher precision arithmetic to the accumulations in the MACs using the lower precision quantized operands \cite{kalamkar2019study, micikevicius-mixed}. The higher precision outcomes from MACs were quantized to a lower precision format to be used for consequent operations. In the following (X + Y) formats, X represents the data format used for MAC operations, and Y represents the arithmetic applied for the accumulations in MAC operations (refer to  \cite{gupta-comparative} for the details for the lower and higher precision arithmetic usage.).      

\textit{a. IFP16 + IFP32:} In 2018, \cite{micikevicius-mixed} noticed that the weights were updated using very small weight gradient values, and applied a lower precision arithmetic IFP16 to the multiplications and a higher precision IFP32 to the accumulations for the weight updates. For example, in the mixed-precision training approach in \cite{micikevicius-mixed}, IFP16 was used to store weights, activations, activation gradients and weight gradients, while IFP32 was used to keep the weight copies for their updates. Along with accumulating IFP16 operands using IFP32 arithmetic, the use of loss scaling allowed the mixed precision training to achieve equivalent accuracy to the IFP32 training while reducing the memory footprint. 

\textit{b. BFloat16 + IFP32:} 
In 2018, the mixed-precision DNN training using (BFloat16 + IFP32) was explored in  \cite{ying2018image}. In 2019, \cite{kalamkar2019study} studied the BFloat16's feasibility for mixed precision training for various DNNs including AlexNet, ResNet, GAN, etc., and concluded that (BFloat16 + IFP32) scheme outperformed the (IFP16 + IFP32) scheme since BFloat16 could represent the same dynamic range of data as IFP32 while using fewer bits.  

\textit{e. FP8 + DLFloat} In 2018, \cite{wang-training} proposed a mixed-precision training method that applies the 5eFP8 format (1 sign-bit, 5-bit exponent, and 2-bit for significand) to the multiplications and DLFloat to the accumulations in MAC operations. The mixed precision method improved resource efficiency such as accuracy per memory footprint and accuracy per memory access, compared to various (FP16 + IFP32) schemes with respect to different FP16 formats. Compared to the previous (FP16 + IFP32) methods, the chunk-accumulation and stochastic rounding schemes were additionally used to minimize the accuracy loss in \cite{wang-training},  The chunk-based accumulation utilized 64 data per chunk instead of one long sequential accumulation to reduce rounding errors. Utilizing stochastic rounding scheme with limited precision format for deep learning was proposed earlier in \cite{gupta}.  
\cite{sun-hybrid8bit} noted that (5eFP8 + DLFloat) training degraded accuracy for DNNs utilizing depth-wise convolutions such as MobileNets. To overcome this issue, \cite{sun-hybrid8bit} proposed to employ two different 8 bit floating-point formats each for forward and backward propagation to minimize the accuracy degradation for compressed DNNs. The mixed-precision training utilized 5eFP8 for backpropagation and another 8-bit floating-point format with (Sign, Exponent, significand) = (1, 4, 3), 4eFP8, for forward propagation. 

\textit{c. DLFloat only:} In 2019, \cite{agrawal-dlfloat} employed the DLFloat for entire training procedure, removing the necessity of data conversions between the multiplications and the accumulations and found that DLFloat could provide better balance between dynamic range and precision than IFP16 and BFloat16 for LSTM networks \cite{Hochreiter-lstm} using Penn Treebank dataset. The DLFloat arithmetic units removed subnormal mode and supported the round-to-nearest up mode to minimize computational complexity. In \cite{agrawal-dlfloat}, the DLFloat arithmetic showed equivalent performance to IFP32  on ResNet-32 using CIFAR10 and ResNet-50 using ImageNet, while using a half of IFP32 bit width.        

\textit{f. INT8-based:} 
Yang et al. \cite{YANG-training} noticed that previously proposed mixed-precision training schemes did not quantize the data in the batch normalization layer, requiring high floating-point arithmetic in some parts of the data paths. To overcome this issue, \cite{YANG-training}  proposed a unified INT8-based quantization framework that quantizes all data paths in DNN including weights, activation, gradient, batch normalization, weight update, etc. into INT8-based data. However, this training method degraded the accuracy to some extent. In 2020, Zhu et al.~\cite{zhu-towards-int8} improved the accuracy, compared to the work of \cite{YANG-training} while keeping unified INT8-based quantization framework. \cite{zhu-towards-int8}  minimized the deviation of the activation gradient direction between before and after quantization by measuring the distance during runtime based on the inner product between the two normalized gradient vectors generated before and after quantization.     

\textit{g. Layer-Wise Adaptive Fixed-Point Training:} In 2020, Zhang et al. \cite{Zhang-fixed-point} proposed a layer-wise adaptive quantization scheme. For example, activation gradient distributions at fully connected layers followed a narrower distribution, requiring more bit-width for the quantizations. \cite{Zhang-fixed-point} quantized AlexNet using INT8 for all the weights and activations and both INT8 (22\%) and INT16 (78\%) for the activation gradients. The quantized AlexNet achieved equivalent accuracy to the one using IFP32 for entire training on ImageNet dataset. 

\subsubsection{Block Floating-Point Training}
Block Floating-Point (BFP) format utilizes a shared exponent for a series of numbers in a data block in order to reduce data-size \cite{wilkinson:algebraic}. Applying BFP to DNNs can improve the resource efficiency in terms of accuracy per memory footprint and per memory access. In addition, BFP utilizes less transistors for simpler adders and multipliers than FP adders and multipliers, resulting in improving accuracy per Joule. Various versions of DNN training methods using BFP were proposed to improve resource efficiency. 

\textit{a. Flexpoint:} 
A DNN-optimized BFP format, \textit{Flexpoint} \cite{koster-flexpoint}, was proposed by Intel, and it was used with the Nervana neural processors. The BFP format used 5 bits for a shared exponent and 16 bits for the significand for the data in a data block. Flexpoint utilized the format of \textit{(Flex N)+M}, where \textit{Flex N} represents variable number of bits for the shared exponent according to the different epochs, and \textit{M} represents the number of bits for the separated significand. For example, the number of exponent bits is adapted based on the dynamic range of the weight values depending on the number of iterations; the dynamic range of the weight values at the current iteration was predicted at the previous iteration. The (\textit{Flex N + 16}) format produced equivalent accuracy to IFP32 in AlexNet using ImageNet dataset and a ResNet using CIFAR-10 dataset, resulting in significant resource efficiency improvement in terms of accuracy per memory footprint and accuracy per memory access.

\textit{b. BFP + FP training:}
Drumond et al. \cite{drumond-training} proposed a hybrid use of BFP and FP for DNN training that uses BFP only for MAC operations and FP for the other operations. Such hybrid training method brought 8.5 $\times$ potential throughput improvement with minor accuracy loss in WideResNet28-10 using CIFAR-100 dataset on a Stratix V FPGA.

\textit{c. Block MiniFloat:} 
\cite{fox-blockminifloat} noticed that ordinary BFP formats were limited in minimizing original data loss with fewer bits and improving arithmetic density per memory access for deep learning applications. To address the two issues, Fox et al. \cite{fox-blockminifloat} proposed the Block Minifloat (BM) along with customized hardware circuit design. The BM<e,m> format follows:
\begin{equation} \label{eq:block-minifloat_format}
y_{out} = \begin{cases} 
      \text{normal mode:} &  (-1)^{sign}\times(1\times2^0 + d_1\times2^{-1}+ ... + d_m\times2^{-m}) \times 2^{exponent - bias - BIAS_{SE}}\\
      \text{subnormal mode:} & (-1)^{sign}\times(d_1\times2^{-1}+ ... + d_m\times2^{-m}) \times 2^{1 - bias - BIAS_{SE}},
   \end{cases}    
\end{equation}
where $bias = 2^{e-1} - 1$ and $BIAS_{SE}$ is a shared exponent value. $BIAS_{SE}$  is scaled according to the maximum value of the data for dot-product operations. For example, BM<2,3> represents a 6-bit data format having 1 sign bit, 2-bit exponent, and 3-bit significand. Such BM variant formats were applied for training. 
Utilizing these 6-bit BM formats produced equivalent accuracy to IFP32 formats but with fewer bits using CIFAR 10 and 100 dataset with ResNet-18, resulting in reduced memory-traffic and low energy consumption. Therefore, BM improved the resource efficiency in term of accuracy per memory access. 

\section{Implementation-Level Resource-Efficient Techniques } \label{sec:implementation}
\begin{figure}
    \centering
\includegraphics[width=4.5in]{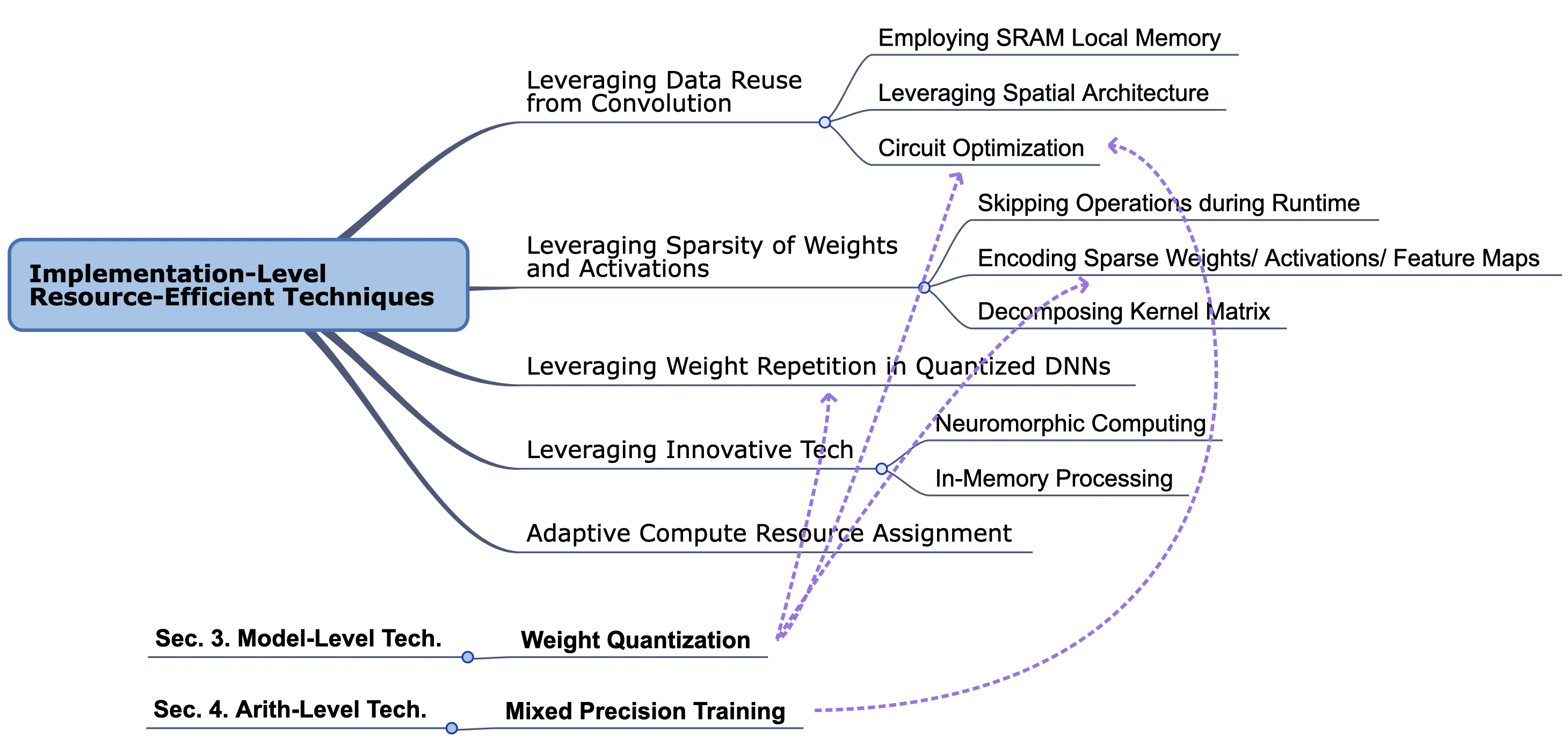}
    \caption{Implementation-level resource-efficient techniques.}
    \label{fig:architectures_class}
\end{figure}

Fig.~\ref{fig:architectures_class} classifies the implementation-level resource-efficient techniques. Most implementation-level techniques have focused on improving energy efficiency and computational speed for MAC operations, since MACs generally occupy more than ~90\% of computational workload for both training and inference tasks in DNNs \cite{sze-efficient}. The implementation-level resource-efficient techniques exploited the characteristics of MACs in DNN including {data reuse}, {sparsity of weights and activations}, and {weight repetition from quantized DNNs}. 

\subsection{Leveraging Data Reuse from Convolution} 
\begin{figure}
    \centering
\includegraphics[width=4.5in]{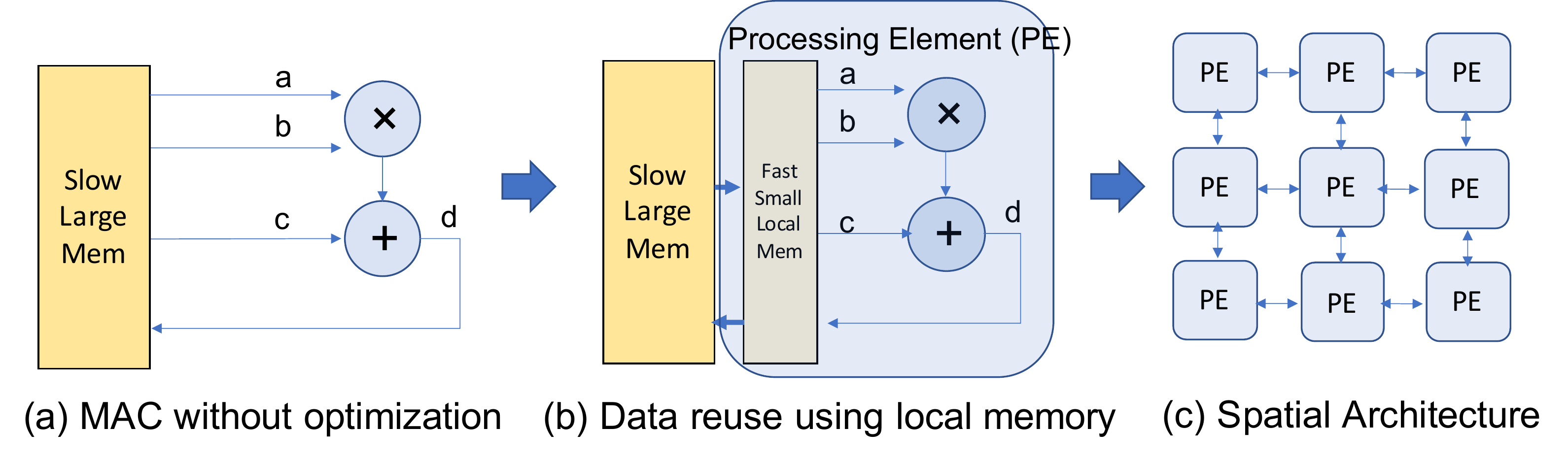}
    \caption{Multiply-and-accumulate dataflow.}
    \label{fig:mac}
\end{figure}

The weights and the activations are heavily reused in convolution operations. For example, the weights of a filter are reused $((H - k_H + 1)\times (W - k_W+1))/stride$ times, where $H=W=4$ (height and width at input channel) and $k_H = k_W = 3$ (height and width for a kernel) in Fig.~\ref{fig:convolution}. Generally, $H$ and $W$ are three orders of magnitude (e.g., 128, 256, etc), $k_H$ and $k_W$ are one order of magnitude (e.g., 3, 5, etc), and $stride$ is either $1$ or $2$. For example, if $H = W= 128$, $k_H = k_W= 3$, and $stride = 1$, each filter is reused $16129$ times for convolutional operations. 
Each input element at a covolutional layer is also reused approximately $ M \times k_{H} \times k_{W}$ times, where $M$ is the number of the total kernels used in the layer. Fig.~\ref{fig:mac} describes the data access patterns for MAC operations used for convolutional layers. In each MAC computation in Fig.~\ref{fig:mac}.(a), the data, \textit{a}, \textit{b}, and \textit{c}, are read from the memory for multiply and add computation, and the result \textit{d} is written back to the memory, where \textit{c} contains a partial sum for the MAC. To save energy consumption, highly reused data for MAC computations can be stored in small local memory as shown in Fig.~\ref{fig:mac}.(b). For example, the power consumption required to access data depends on where the data are located – accessing data from off-chip memory, DRAM, generally requires two orders of magnitude more than from on-chip memory \cite{chen-eyeriss}.  
For commercially available CPUs or GPUs, transforming convolutional operations into matrix multiplications can leverage such data reuse properties to accelerate the convolution operations by utilizing highly optimized BLAS libraries \cite{vanhoucke-improving}. Many research works presented how to leverage such data reuse properties to improve the resource efficiency. 

\subsubsection{Employing SRAM Local Memory near to Processing Elements:}
The use of SRAM buffers reduces the energy consumed by DNNs by up to two orders of magnitude, compared to DRAM. Similar to Fig.~\ref{fig:mac}.(b), \textit{Dianao} architecture~\cite{chen-dianao} employed one Neural Functional Unit (NFU) integrated with three separated local buffers, each for holding $16$ input neurons, $16 \times 16$ weights, and $16$ output neurons, in order to optimize circuitry for MAC operations. The weights and the activations stored to the local memories were reused efficiently by additionally using internal registers to store the partial sums and the circular buffer. The NFU is a three-stage pipelined architecture consisting of the multiplication, the adder-tree, and the activation stage. In the multiplication stage, $256$ multipliers support the multiplications based on the weight connections between $16$ input and $16$ output neurons. In the adder-tree stage, $16$ feature maps are generated from the multiplications based on adder-tree structure. In the activation stage, the $16$ feature maps are approximated for the $16$ activations by using piece-wise linear function approximation. DianNao with $65 nm$ ASIC layout brought up to $118 \times$ speedup and reduced the energy consumption by $21 \times$, compared to a 128-bit 2GHz SIMD processor over the benchmarks used in \cite{chen-dianao}. One of the following studies adapted DianNao to deploy it on a supercomputer and named it as DaDianNao~\cite{chen-dadiano}. Since the number of weights is generally larger than the number of input activations for convolution operations, {DaDianNao} stored a big chunk of weights and shared them to multiple NFUs by using \textit{a central embedded DRAM} to reduce the data movement cost in delivering the weights associated to each NFU. 

\subsubsection{Leveraging Spatial Architecture:}
Designing PEs and their local memory according to data reuse properties of MAC operations improved energy efficiency on FPGAs and ASIC \cite{chen-eyeriss, chen-eyeriss2}. For example, Google TPU employs a systolic array architecture to send the data directly to an adjacent Processing Element (PE) as shown in Fig.~\ref{fig:mac}.(c) \cite{tpu_google}. Chen et al.~\cite{chen-eyeriss} noticed that the computational throughput and energy consumption of CNNs mainly depended on data movement rather than computation and proposed a ``row-stationary'' spatial architecture (a variant of Fig.~\ref{fig:mac}.(c)), \textit{Eyeriss}, to support parallel processing with minimal data movement energy cost by fully exploiting the data reuse property. 
For example, the three PEs in the first column in Fig~\ref{fig:mac}.(c) can be assigned to compute the first row of the convolution output using a $3\times3$ filter - the three elements on each row of the kernel are stored to the local memory on each PE (i.e., ``row-stationary'' structure in \cite{sze-efficient}), and all the elements in the kernel are reused during convolution, generating the first row of the output. In this case, the partial summation values are stored back to the local memory on each PE.

\subsubsection{Circuit Optimization:}
Exploring binary weights~\cite{courbariaux-binnaryconnect} with binary inputs offered the opportunity to explore XNOR gates for the efficient implementation of CNN~\cite{rastegari-xnor-net}, improving the accuracy per memory foot print and per Joule. In 2021, \cite{zhao-cambricon-q} proposed hardware-friendly statistic-based quantization units and near data processing engines to accelerate mixed precision training schemes by minimizing the number of accesses to higher precision data.     

\subsection{Leveraging Sparsity of Weights and Activations} 
In the forward pass, negative feature map values are converted to zeros after ReLU activation functions, making the activation data structure sparse. In addition, the trained weight values follow a sharp Gaussian distribution centered at zero, locating most of the weights near to zero. Quantizing such weights makes the weight data structure sparse, so the sparse weights can be fully exploited on the quantized networks such as binarized DNNs \cite{courbariaux-binarized, courbariaux-binnaryconnect} and ternary weight DNNs \cite{li-ternary, zhu-ternary}.  

\subsubsection{Skipping Operations during Runtime:} \label{sec:skipping operations} In 2016, several methods to conditionally skip MAC operations were proposed simultaneously \cite{chen-eyeriss, albericio-cnvlutin, liu-cambricon}. {Eyeriss}~\cite{chen-eyeriss} employed clock gating to block the convolution operations during runtime when either the weight or the activation was detected as zero in order to save computational power. \textit{Cnvlutin}~\cite{albericio-cnvlutin} skipped  MAC operations associated with zero activations by employing separated ``neuron lanes'' according to different channels.
Similarly, \cite{liu-cambricon} proposed \textit{Cambricon-X} that fetches the activations associated with any non-zero weights for convolutions by using the ``step indexing'' in order to skip the MAC operations associated with the zero weights. 
\textit{Cambricon-X }brought 6 $\times$ resource-efficiency improvement in terms of accuracy per Joule, compared to the original \textit{DianNao} architecture.    
In 2017, Kim et al.~\cite{kim-zena} proposed \textit{ZeNa }that performs MAC operations only if both the weights and the activations are non-zero values. 
In 2018, Akhlaghi et al.~\cite{akhlaghi-snapea} proposed a runtime technique, \textit{SnaPEA}, that performs MAC operations associated with positive weights first and then negative weights later while monitoring the sign of the partial sum value. Since the activation values from ReLU are always greater or equal to zero, the convolution operation can be terminated once the partial sum value becomes negative. Notice that such decision should be performed during runtime, since the zero valued activation patterns depend on the test images.
In 2021, another method skipping zero operations,\textit{ GoSPA} \cite{deng-gospa}, was proposed, which is similar to ZeNa in that MAC operations were performed only when both input activations and weights were non-zero values. \cite{deng-gospa} constructed ``Static Sparsity Filter" module by leveraging the property that the weight values are static while the activation values are dynamic to filter out zero activations associated with non-zero weights on the fly before MAC operations. Such skipping operation optimization techniques improved the accuracy per Joule, since the transistors associated with skipped operations were not toggled during runtime, saving dynamic power consumption.  

\subsubsection{Encoding Sparse Weights/Activations/Feature Maps:} 
Since memory access operations dominate the power consumption in deep learning applications, fetching the weights less frequently from memory by encoding and compressing the weights and the activations can improve the resource efficiency such as the accuracy per memory footprint, per memory access, and per Joule. 
Han et al.~\cite{han-deepcompression, han-eie} utilized the Huffman encoding scheme to compress the weights. The quantized DNN reduced the memory footprint of AlexNet on ImageNet dataset by $35$ times without losing accuracy. In \cite{han-deepcompression, han-eie}, a three stage pipelined operation was performed in order to reduce the memory footprint of DNNs as follows. The pruned sparse quantized weights were stored with Compressed Sparse Row (CSR) format and then divided into several groups. The weights in the same group were shared with the average value over the weights in the group, and they were re-trained thereafter. Huffman coding was used to compress the weights further. Parashar et al.~\cite{parashar-scnn} employed an encoding scheme to compress sparse weights and activations and designed an associated dataflow, "Compressed-Sparse Convolutional Neural Networks (SCNN)", to minimise data transfer and reduce memory footprint.
Aimar et al.~\cite{aimar-nullhop} proposed \textit{NullHop }that encodes the sparse feature maps by using two sequentially ordered (i.e., internally indexed) additional storage, one for a 3D mask to indicate the positions of non-zero values and the other for storing the non-zero data sequentially in the feature map. For example, `0's are marked at the position of zero values in the 3D mask, otherwise `1's are marked. Decoding refers to both the 3D mask and the non-zero value list. Rhu et al. \cite{rhu-compressing} presented \textit{HashedNet} that utilizes a low cost hash function to compress sparse activations. 
The virtualized DNN (vDNN) \cite{rhu-vdnn} compressed sparse activation units using the ``zero-value compression'' technique to minimize the data transfer cost between GPU and CPU. The vDNN allowed users to utilize both GPU and CPU memory for DNN training. 

\subsubsection{Decomposing Kernel Matrix:} Li et al.~\cite{li-squeezeflow} proposed SqueezeFlow that reduces the number of operations for convolutions by decomposing the kernel matrix into non-zero valued sub-matrices and zero-valued sub-matrices. This method can improve the accuracy per Joule. 

\subsection{Leveraging Weight Repetition in Quantized DNNs}
Hedge et al.~\cite{hedge-ucnn} noticed that the identical weight values were often repeated in quantized DNNs such as binary weight DNNs \cite{courbariaux-binarized, courbariaux-binnaryconnect} and ternary weight DNNs \cite{li-ternary, zhu-ternary} and proposed the Unique Weight CNN Accelerator (UCNN) that reduces the number of memory accesses and the number of operations by leveraging the repeated weight values in the quantized DNNs. For example, if a $2\times2$ kernel consisting of $\{k_{1,1}, k_{1,2}, k_{2,1}, k_{2,2}\}$ performs a convolution with the activation maps, $\{a_{1,1}, a_{1,2}, a_{2,1}, a_{2,2}\}$. The conventional covolutional operation, $\Sigma_{i=1,j=1}^{i=2,j=2} k_{i,j}\times a_{i,j}$, requires $8$ read memory accesses, $4$ multiplications, and $3$ additions. If two of the weights in the kernel are identical (e.g., $k_{1,1} = k_{2,2}$ and $k_{1,2} = k_{2,1}$), the covolutional operation can be performed using: $k_{1,1}(a_{1,1}+a_{2,2}) + k_{1,2}(a_{1,2}+a_{2,1})$, requiring $6$ read memory accesses, $2$ multiplications, and $3$ additions. The UCNN improved the accuracy per Joule by $1.2-4\times$ in AlexNet and LeNet on Eyeriss architecture using ImageNet dataset.   

\subsection{Leveraging Innovative Technology}

Many research attempts have leveraged innovative computing architecture technologies such as neuromorphic computing and in-memory processing as follows. 

\subsubsection{Neuromorphic Computing:}
Neuromorphic computing mimics the brain, including brain components such as neurons and synapses; furthermore, biological neural systems include axons, dendrites, glial cells, and spiking signal transferring mechanisms \cite{1333071}. The memristor, `memory resistor', is one of the most widely used devices in neuromorphic computing. \textit{ISAAC}~\cite{shafiee-isaac} replaced MAC operation units with the memristor crossbars based on \textit{DaDianNao} architecture. In the crossbars, every wire in horizontal wire array was connected to every wire in vertical wire array with a resistor. Different level voltages, $V = [v_1, v_2, ..., v_m]$, were applied to the horizontal wires connected to a vertical wire by the different resistors, $R = [1/g_1, 1/g_2, ..., 1/g_m]$. With mapping $v_i$ to input elements and $g_i$ to weights, where $i = 1, ..., m$, the output current, $I$, from the vertical wire can be represented as MAC operations in a layer, $I = \Sigma_i^m (v_i \times g_i)$, based on the Kirchhoff's law. Multiple MAC operations can be performed by collecting the currents from multiple vertical wires.  \textit{ISAAC} employed digital-to-analog converters to receive the input elements and covert them into the appropriate voltages and analog-to-digital converters to convert the current values into digitized feature map values. Due to lack of re-programability of resistors in the crossbars, {ISAAC }architecture was only available for the inference tasks. {ISAAC }improved $5.5 \times$ accuracy per Joule, compared to full-fledged \textit{DaDianNao}. As another neuromorphic computing approaches, many research attempts implemented Hodgkin-Huxley and Morris Lecar models \cite{https://doi.org/10.1113/jphysiol.1952.sp004764} that describes the activity of neurons using nonlinear differential equations in hardware simulators \cite{10.1007/s10470-012-9888-6,764940,4541730,788662,775396,DBLP:conf/nips/SimoniCSCD00,1262112,7146404,4541446,5951803,493808}. 
Several studies implemented neuromorphic architectures in ASIC, including TrueNorth~\cite{7229264}, SpiNNaker~\cite{6515159}, Neurogrid~\cite{6805187}, BrainScaleS~\cite{6272131}, and IFAT~\cite{6981816}. Please refer to \cite{schuman2017survey} for a comprehensive survey of neuromorphic computing.  

\subsubsection{In-Memory Processing:} Scaling down the size of transistors enables energy efficiency and high performance for Von-Neumann computing systems. However, it became very challenging in the era of sub-10nm technologies due to physical limitations~\cite{8675237,10.1145/977091.977115,8710283}. To address this challenge, researchers proposed a paradigm of in-memory processing to improve performance and energy efficiency by integrating computations units into memory devices~\cite{10.1145/3445814.3446749,9065462,8686604}. Several studies proposed to enable in-memory processing to accelerate DNNs ~\cite{8870210,8465866,9251855,7551408,10.1145/3093337.3037702, yin-xnor-sram}. For example, XNOR-SRAM~\cite{yin-xnor-sram} integrated the XNOR gates and accumulation logic into SRAM to fetch data from SRAM and perform MAC operation in one cycle. Notice that this approach was applicable for binarized DNNs such as \cite{courbariaux-binarized}.

\subsection{Adaptive Compute Resource Assignment} 
This subsection comprises the methods assigning runtime compute resources adaptively to the DNN inference workload to improve resource efficiency. The implementation of the DNNs can be adapted to the accuracy requirements of the applications by using various runtime implementation techniques as follows.     

\subsubsection{Early Exiting:}
The required depth of DNN depends on the problem complexity. The ``early exiting'' technique allows a DNN to classify an object as early as possible by having multiple exit classifier points in a single DNN \cite{panda-conditional, teer-branchy, teer-distributed}.
The early exiting technique was applied to distributed computing systems, addressing concern about privacy, response time, and higher quality of experience~\cite{teer-distributed}. Such early exiting methods minimized the compute resources and the inference latency, improving the accuracy per Joule, per operation, and per core utilization. Please refer to \cite{matsubara2021split} for the details on the early exiting techniques. 

\subsubsection{Runtime Channel Width Adaptation:}
The runtime channel width adaptation pruned unimportant filters during runtime. In 2018, Fang et. al \cite{fang-nestdnn} presented a single DNN model, \textit{NestDNN}, being able to switch between multiple capacities of the DNN during runtime according to the accuracy and inference latency requirement. During training, unimportant filters from the original model were pruned to generate the smallest possible model, ``\textit{seed model}''. Each re-training, some of pruned filters were added to the seed model while fixing the filter parameters from the previous training. Since the seed model was descended from the original model, the accuracy for each capacity in NestDNN was higher than the model having the identical capacity trained from the scratch. Similarly, Yu et. al \cite{yu-slimmable} proposed another runtime switchable DNN model,\textit{ Slimmable Neural Network}, in which a larger capacity model shared the filter parameters from a smaller capacity model. 

\subsubsection{Runtime Model Switching:}
Lee et al. \cite{lee-tod} selected the best performing object detectors between multiple DNN detectors during runtime according to dynamic video content to improve the accuracy per core utilization and per Joule. 
Lou et al. \cite{lou-dynamic-ofa} switched between multiple DNNs, generated from the Once-For-All NAS of \cite{cai-ofa}, during runtime according to dynamic workload. For example, when the inference latency of a DNN was increased due to a newly assigned workload, a runtime decision maker downgraded the current DNN during runtime to meet a latency constraint.   
Such runtime model switching approaches were appropriate when memory resources were sufficient, since the multiple DNNs should be pre-loaded in DRAM.

\section{Interrelated Influences}  \label{sec:influence}
This section discusses the influence from higher- to lower-level techniques as shown in Fig.~\ref{fig:survey-diagram}. 

\subsection{Influences of Model-Level Techniques on Arithmetic-Level Techniques}
Weight quantization \cite{courbariaux-binnaryconnect, li-ternary, zhu-ternary} in model-level techniques influenced arithmetic-level techniques as shown in Fig.~\ref{fig:arithmetic_level_diagram}. The multiplications using the quantized binary weights can be replaced with multiplexers, removing multiplication arithmetic operations. 
The resource efficiency from the model-level techniques can be further improved by utilizing the arithmetic-level techniques.
For example, quantized DNNs such as ternary weight and binarized DNNs allowed INT8 arithmetic to be used in training \cite{wu2018training, YANG-training}. 
When reduced precision DNNs suffered from zero gradients, the reduced precision arithmetic was replaced with a hybrid version arithmetic using both BFP and FP \cite{drumond-training} or the Block MiniFloat format \cite{fox-blockminifloat}.

\subsection{Influences of Model-Level Techniques on Implementation-Level Techniques}
Weight quantization in model-level techniques influenced the implementation-level techniques as shown in Fig.~\ref{fig:architectures_class}. Pruning weights can bring sparsity in the hardware architecture while pruning filters \cite{luo-thinet, li-pruning} maintains dense structure. 
Weight quantization in the model-level techniques allows a DNN to utilize fewer bits for weights in order to save memory resource usage, requiring customized hardware. For example, EIE~\cite{han-eie} is an inference accelerator with weights quantized by 4 bits. To implement the weight quantization method effectively, EIE utilized weight sharing to reduce the model size further and fit the compressed DNN into the on-chip SRAM. 
Exploring binary weights~\cite{courbariaux-binarized} with binary inputs offered the opportunity to explore XNOR gates for the efficient implementation of CNNs~\cite{rastegari-xnor-net}, improving the accuracy per memory footprint and per Joule.  
In \cite{tridgell-unrolling}, ternary neural networks \cite{li-ternary, zhu-ternary} were implemented on FPGAs by unrolling convolution operations. 
Since quantized DNNs \cite{li-ternary, zhu-ternary} increased the number of repeated weights in DNNs, UCNN~\cite{hedge-ucnn} leveraged the property of the repeated weight values in quantized DNNs to improve resource efficiency such as accuracy per memory access and per operation. As main limitation, the weight quantization methods such as \cite{courbariaux-binnaryconnect, zhu-ternary, li-ternary} were not suitable for commercially available CPUs and GPUs, since such computing platforms do not support binary and ternary weights in hardware. Therefore, the implementation of weight quantization methods on CPUs or GPUs might not improve accuracy per Joule as higher precision arithmetic still was required in part of data path in training and inference. The bottleneck structures generated by compact convolutions in \cite{sandler-mobilenetv2, iandola-squeezenet} can be used to reduce the data size transferred between a local device and an edge server for the efficient implementation of edge-based AI \cite{matsubara2021split}.

\subsection{Influences of Arithmetic-Level Techniques on Implementation-Level Techniques}
The arithmetic-level techniques influenced the implementation-level techniques as follows. 

First, the research in arithmetic utilization acted as a catalyst for commodity CPUs and GPUs. For example, the mixed precision research \cite{micikevicius-mixed} laid a foundation for tensor cores in latest NVIDIA GPUs, which can accelerate the performance of deep learning workloads by supporting a fused multiply–add operation and the mixed precision training capability in hardware~\cite{blanchard2020mixed}. The BFloat16 format~\cite{burgess-bfloat16} designed by Google overcomes the limited accuracy issue of the IFP16 format by providing the same dynamic range as IFP32, and it is supported in hardware in Intel Cooper Lake Xeon processors, NVIDIA A100 GPUs, and Google TPUs. In 2016, NVIDIA Pascal GPUs supported IFP16 arithmetic in hardware to accelerate DNN applications. In 2017, NVIDIA Volta GPUs supported IFP16 tensor cores. In 2020, the NVIDIA Ampere architecture supported tensor cores, TF32, BFloat16, and sparsity acceleration in hardware to accelerate MACs~\cite{nvidia_ampere}. The Graphcore company developed the Intelligent Processing Unit (IPU), which employs local memory assigned to each processing unit with support for a large number of independently operating hardware threads~\cite{jia2019dissecting}. The IPU is an efficient computing architecture customized to ``fine-grained, irregular computation that exhibits irregular data access''. 

Secondly, the arithmetic-level techniques led to specialized custom accelerators for deep learning. There is ample evidence in the arithmetic-level literature such as \cite{fox-blockminifloat, drumond-training, Bordawekar-abali, wang-training, Yang-designing} that even smaller operators (e.g., 16 bits or even less) have almost no impact on the accuracy of DNNs. For example, DianNao~\cite{chen-dianao} and DaDianNao~\cite{chen-dadiano} were customized to 16-bit fixed-point arithmetic operators instead of word-size (e.g., 32-bit) floating-point operators. ISAAC~\cite{shafiee-isaac} is a fully-fledged crossbar-based CNN accelerator architecture, which implemented a memristor-based logic based on resistive memory, suitable for 16-bit arithmetic for DNN workloads. Wang et al.~\cite{wang-training} designed their customized 8-bit floating point arithmetic multiplications with 16-bit accumulations on an ASIC-based hardware platform with a $14 nm$ silicon technology to support energy-efficient deep learning training. The Eyeriss~\cite{chen-eyeriss} and SnaPEA~\cite{akhlaghi-snapea} accelerators were customized to 16-bit arithmetic. UCNN~\cite{hedge-ucnn} utilized 8-bit and 16-bit fixed point configurations. SCNN~\cite{parashar-scnn} utilized 16-bit multiplication and 24-bit accumulation.

Lastly, the mixed precision training schemes were accelerated in hardware by minimizing the data conversion overhead between lower and higher precision formats in updating weights and activations \cite{zhao-cambricon-q}. Also, the stochastic rounding scheme was supported in hardware in Intel Loihi processor \cite{davies-lohi} and Graphcore IPU \cite{jia2019dissecting}, since it was often required for quantizing weights and activations during training \cite{wu2018training, gupta, YANG-training}. 

\section{Future Trend for Resource-Efficient Deep Learning} \label{sec:open issues}
Open research issues in resource-efficient deep learning emerge in an attempt to improve the resource efficiency further, compared to the state-of-the-art resource-efficient techniques discussed in this paper.  

\subsection{Future Trend for Model-Level Resource-Efficient Techniques}
Recently, edge-based computing has become pervasive, and fitting DNN models into such resource-constrained devices for inference tasks has become extremely challenging. 

\subsubsection{Improving Physical Resource Efficiency under Very Low Compute Resource Budget:}
Many researchers considered keeping dense network structures after pruning parameters, including pruning channels \cite{liu-learning, gao-dynamic}, filters \cite{li-pruning, luo-thinet}, etc., to implement the pruned networks efficiently on commercially available CPUs and GPUs. Since then, various budget-aware network pruning methods were proposed, given a resource budget such as the number of floating point operations \cite{gordon2018morphnet} and the number of neurons \cite{lemaire2019structured} for the inference task. \textit{NetAdapt}~\cite{yang-netadapt} pruned the filters as it measured physical resources such as latency, energy, memory footprint, etc. to improve the physical resource efficiency directly rather than abstract resource efficiency. Along with the fast technology development in computer networks and wireless communications, research attempts to improve physical resource efficiency are expected to continue to deploy appropriate DNN models on extremely low resource devices such as mobile, IoT, and edge devices.   

\subsubsection{Neural Network Search Methods Combined with Domain Specific Knowledge:}
In 2016 and 2017, handcrafted compressed DNNs were presented such as SqueezeNet \cite{iandola-squeezenet}, MobileNet \cite{howard-mobilenets}, ShuffleNet \cite{zhang-shufflenet}, and DenseNet \cite{huang-densely}, and they improved both abstract and physical resource efficiency. 
Various NAS methods \cite{tan-efficientnet, tan-efficientdet, Tan-mnasnet, he-amc} assisted to seek the optimized DNN models (e.g., least sufficient models) by searching candidate spaces according to the training dataset, and the compressed models found by the NAS methods generally showed superior physical resource efficiency to the handcrafted compressed DNNs. As mobile and edge devices become prevalent, we expect that automatic search methods integrating with domain specific model compression methods are expected to be paid attention in the future. For example, performance-aware NAS methods for resource-constrained devices have been vividly paid attention \cite{anderson-performance, Tan-mnasnet, tan-efficientdet, tan-efficientnet, yang-netadapt} since 2019. Such performance-aware NAS methods can be enhanced by adopting recent domain specific model-level resource-efficient techniques such as \cite{han-deepcompression, iandola-squeezenet, howard-mobilenets, li-pruning, frankle-lottery}. 

\subsubsection{Theoretical Studies Behind Model-Level Resource-Efficient Techniques:}
The bias-variance trade-off~\cite{geman} is behind the model-level resource-efficient techniques. For example, compressed models having fewer parameters increase the regularization effect on the accuracy, minimizing overfitting issues \cite{han-deepcompression, geman}. For example, \cite{frankle-lottery} proposed the lottery ticket hypothesis in that better (or equivalent) performing sub-DNNs using fewer weights exist inside a dense, randomly-initialized, feed-forward DNN.  In order to seek the better performing sparse sub-DNNs, ``winning tickets'', the survived weights from weight pruning were re-trained by replacing the survived weights with the random weights initially used to train the original dense DNN. The lottery ticket hypothesis implies that such sparse DNNs could be found in even compressed dense DNNs. We expect that such theoretical studies supporting model-level resource-efficient techniques can be paid attention in the future.  

\subsection{Future Trend for Arithmetic-Level Resource-Efficient Techniques} 
As edge- and mobile-based devices becomes pervasive for AI applications, open research issues emerge in the attempts to improve further physical resource efficiency on such resource-constrained devices, compared to state-of-the-art arithmetic-level techniques.   

\subsubsection{Adapting Arithmetic Precision Level to Numerical Properties of DNNs}
In 2011, Vanhoucke et al. \cite{vanhoucke-improving} demonstrated the feasibility of INT8 arithmetic for inference tasks using a shallow depth neural network on an Intel x86 architecture. In 2015, Gupta et al. \cite{gupta} demonstrated that employing FiP16 with a stochastic rounding scheme for a shallow depth neural network produced equivalent accuracy using MNIST and CIFAR10 to that using IFP32. In 2018, \cite{micikevicius-mixed} presented the guidelines for mixed precision training. The guidelines contained the information on how to deploy different-level arithmetic precision on different computing components in MAC operations. The guidelines led to  further research attempts in hardware optimization for mixed precision training \cite{zhao-cambricon-q}. We expect that the research attempts in adapting an arithmetic precision to DNN computing components according to their numerical stability characteristics will continue in the future.  

\subsubsection{Adapting Arithmetic Format to Problem Complexity:}
Since floating point arithmetic is computationally intensive, several studies have removed floating point arithmetic in training tasks. For example, Wu et al.~\cite{wu2018training} demonstrated that quantized networks such as binary or ternary weight networks can be trained using INT8 arithmetic along with a scaling and a stochastic rounding scheme. 
In 2020, Yang et al.~\cite{YANG-training} demonstrated that quantizing  weights and activations with INT8 format while applying INT24 arithmetic to the weight updates could accelerate training and inference tasks for various ResNet models using ImageNet with minor accuracy loss, compared to those using IFP32. Recently, RNS-based quantization was applied to various DNNs \cite{samimi-res-dnn, salamat-rnsnet}. It will continue in the future to explore how to adapt a number format to given DNN structures for inference and training tasks in order to improve resource efficiency further on resource-constrained devices.    

\subsection{Future Trend for Implementation-Level Resource-Efficient Techniques} 
In general, there are two ways to accelerate DNN computations. One is to optimize DNN computations on given compute architecture such as CPUs and GPUs. The other is to customize dataflow on FPGAs and ASIC.       

\subsubsection{Leveraging Spatial and Temporal Data Access Pattern with Lower Precision Arithmetic on CPUs and GPUs:}
A decade ago, \cite{vandeGeijn2011} accelerated DNN computations on a SIMD CPU by leveraging the data reuse property from MAC operations using SSE instructions and fast fixed point arithmetic. NVIDIA Tensor cores and Google TPUs support a customized arithmetic precision format such as IFP16, BFloat16, etc. and customized datapath in hardware for deep learning applications \cite{nvidia_ampere, tpu_google}. The research attempts to leverage spatial and temporal data access patterns with lower precision arithmetic in commercially available CPUs and GPUs will continue in the future.   

\subsubsection{Leveraging Spatial and Temporal Data Access Pattern with Lower Precision Arithmetic on FPGAs and ASIC:}
In 2014, \cite{chen-dianao} stressed the limitation of commercially available GPUs and CPUs for DNN applications: ``While a cache is an excellent storage structure for a general-purpose processor, it is a sub-optimal way to exploit reuse because of the cache access overhead (tag check, associativity, line size, speculative read, etc.) and cache conflicts.'' To overcome this limitation, \cite{chen-dianao} proposed a SIMD style hardware accelerator, \textit{DianNao}, that employs three separate local on-chip memories (SRAMs) to maximize the performance by fully leveraging the data reuse property. \cite{chen-dadiano} pointed out that \textit{DianNao} was still limited in the memory bandwidth to access massive weights in covolutional layers and proposed \textit{DaDianNao} architecture employing large eDRAMs with four banks to store and share the weights in the eDRAMs efficiently. In 2016, \cite{chen-eyeriss} pointed out that the data movement cost is still dominant, compared to the computation cost for DNN applications on the SIMD/SIMT architectures of \cite{chen-dadiano, chen-dianao} and proposed a dataflow architecture to minimize energy consumption caused by data movement in DNN applications. Since 2016, most implementation-level techniques leveraged the sparsity of weights and activations in DNNs to minimize the number of arithmetic operations during runtime \cite{albericio-cnvlutin, liu-cambricon, kim-zena, akhlaghi-snapea, deng-gospa} and the data transfer cost required to store and transfer the sparse weights and activations \cite{han-eie, aimar-nullhop, parashar-scnn, rhu-compressing, rhu-vdnn}. The research efforts to customize DNN dataflow by leveraging spatial and temporal data access patterns with lower precision arithmetic are expected to continue in the future.   

\subsubsection{Resource-Efficient Implementation on Distributed AI Compute Platforms:}
Resource-efficient techniques on distributed AI such as split federated learning \cite{ha-spatio-splitlearning, thapa2021splitfed} and early exiting \cite{teer-branchy, teer-distributed} have recently attracted a great deal of attention thanks to fast wireless network technology development. The main open research issues include data communication overhead between an edge device and the cloud and energy consumption, required to run DNNs on a lower power (or battery) edge device. For example, many research attempts leveraged the bottleneck structure of a DNN to save the data communication bandwidth, but such attempts could degrade the accuracy significantly in the DNNs employing compact convolutions \cite{matsubara2021split}. Thus, adapting such model compression techniques to distributed AI environments can be paid attention in the future in order to save energy consumption on edge devices and the bandwidth required for communication between an edge device and a cloud. For example, encoding and decoding offloading data from the cloud to edge devices or vice versa can minimize the data communication overhead \cite{yao-deepcompressive}. Such resource-efficient encoding/decoding schemes for split learning (or inference) tasks can draw attention in the future.

\subsubsection{Neuromorphic Computing for Deep Learning:} Neuromorphic computing can lead to dramatic changes in energy efficiency for deep learning \cite{schuman2017survey}. This expectation is based on the fact that neuromorphic computing is not an incremental improvement of existing von Neumann architectures requiring considerable energy due to substantial instruction fetch/decode operations, but an fully optimized dataflow optimization customized to the activity of a neural network. Therefore, neuromorphic computing research can be paid attention in the future to maximize the accuracy per Joule.

\section{Conclusion} \label{sec:conclusions}
Our survey is the first to provide a comprehensive survey coverage of the recent resource-efficient deep learning techniques \textit{based on the three-level hierarchy including model-, arithmetic-, and implementation-level techniques}. Our survey also utilizes multiple resource efficiency metrics to clarify which resource efficiency metrics each technique can improve. 
For example, most model-level resource-efficient techniques contribute to improving abstract resource efficiency,  
while the arithmetic- and the implementation-level techniques directly contribute to improving physical resource efficiency by employing reduced precision arithmetic and/or optimizing the dataflow of DNN architectures. Therefore, the efficient implementation of the model-level techniques on given compute platforms is essential to improve physical resource efficiency~\cite{sze-efficient}. 

In the future, we expect that the three-level resource-efficient deep learning techniques can be adapted to distributed AI applications, along with fast wireless communication technology development. Since edge or mobile devices are subjected to physical resource constraints such as power, memory, and inference speed, the implementation should consider such constraints for the distributed AI applications. The state-of-the-art works include the NAS variants of \cite{Tan-mnasnet, florian-constrained, umar-leveraging} that seek the optimal performing DNN models fitted to the resource-constrained edge-devices. Improving such NAS variants by combining them with various model-, arithmetic-, and implementation-level resource-efficient techniques can be paid attention in the future. 

Finally, our survey suggests that the bias-variance trade-off~\cite{geman} is behind the model-level resource-efficient techniques. According to the trade-off, a DNN having fewer parameters increases the regularization effect on the accuracy, minimizing overfitting issues. Therefore, there exists the least sufficient model size that produces the best accuracy on test dataset according to the problem complexity and the training data quantity and quality. Similarly, \cite{frankle-lottery} claimed the lottery ticket hypothesis in that better (or equivalent) performing sub-DNNs using fewer weights exist inside a dense feed-forward DNN. Einstein quoted ``Everything should be made as simple as possible, but not simpler.'' We hope that our survey will contribute to machine learning, arithmetic, and system community by providing them with a comprehensive survey for various resource-efficient deep learning techniques as guidelines to seek DNN structures using least sufficient parameters and least sufficient precision arithmetic on particular compute platforms, customized to the problem complexity and the training data quantity and quality.       

\textbf{Acknowledgments}
\\This project has received funding by the Engineering and Physical Sciences Research Council under the grant agreement No. EP/T022345/1 and by CHIST-ERA under the grant agreement No. CHIST-ERA-18-SDCDN-002 (DiPET). This research was also partially supported by National R\&D Program through the National Research Foundation of Korea (NRF) funded by Ministry of Science and ICT (2021M3H2A1038042)

\bibliographystyle{plain}
\bibliography{references}

\end{document}